\documentclass[runningheads]{llncs}

\usepackage{eccv}
\usepackage{eccvabbrv}

\usepackage{graphicx}
\usepackage{booktabs}

\usepackage[accsupp]{axessibility}

\usepackage[pagebackref,breaklinks,colorlinks,citecolor=eccvblue]{hyperref}

\usepackage{orcidlink}

\usepackage{amsmath}
\DeclareMathOperator*{\argmax}{arg\,max}
\DeclareMathOperator*{\argmin}{arg\,min}
\usepackage{float}

\usepackage{multirow}
\usepackage{adjustbox}
\usepackage{appendix}

\usepackage{pifont}
\newcommand{\cmark}{\ding{51}}
\newcommand{\xmark}{\ding{55}}

\usepackage[dvipsnames]{xcolor}
\usepackage{soul}

\newcommand{\mr}[2]{\multirow{#1}{*}{#2}}

\definecolor{cadmiumgreen}{rgb}{0.0, 0.42, 0.24}
\definecolor{darkpastelgreen}{rgb}{0.01, 0.75, 0.24}
\definecolor{brightpink}{rgb}{1.0, 0.0, 0.5}

\begin{document}

\title{On the Utility of 3D Hand Poses for
Action~Recognition}

\titlerunning{HandFormer}

\author{Md Salman Shamil \inst{1}\orcidlink{0000-0001-5556-1145} \and
Dibyadip Chatterjee \inst{1}\orcidlink{0000-0002-2651-3045} \and
Fadime Sener \inst{2}\orcidlink{0000-0001-5004-6005}\and \\
Shugao Ma \inst{2}\orcidlink{0000-0002-4986-2221}\and
Angela Yao \inst{1}\orcidlink{0000-0001-7418-6141}
}

\authorrunning{M.S.~Shamil et al.}
\institute{National University of Singapore \and
Meta Reality Labs \\
\email{\{salman, dibyadip, ayao\}@comp.nus.edu.sg} \\ \email{\{famesener, shugao\}@meta.com}\\
{\small \color{brightpink}\url{https://s-shamil.github.io/HandFormer/}}}

\maketitle
\begin{abstract}
3D hand pose is an underexplored modality for action recognition. Poses are compact yet informative and can greatly benefit applications with limited compute budgets. However, poses alone offer an incomplete understanding of actions, as they cannot fully capture objects and environments with which humans interact. We propose HandFormer, a novel multimodal transformer, to efficiently model hand-object interactions. HandFormer combines 3D hand poses at a high temporal resolution for fine-grained motion modeling with sparsely sampled RGB frames for encoding scene semantics. Observing the unique characteristics of hand poses, we temporally factorize hand modeling and represent each joint by its short-term trajectories. This factorized pose representation combined with sparse RGB samples is remarkably efficient and highly accurate. Unimodal HandFormer with only hand poses outperforms existing skeleton-based methods at 5$\times$ fewer FLOPs. With RGB, we achieve new state-of-the-art performance on Assembly101 and H2O with significant improvements in egocentric action recognition.


\end{abstract}

\section{Introduction}
\label{sec:intro}

The popularity of AR/VR headsets has driven interest in recognizing hand-object interactions, particularly through egocentric~\cite{damen2018scaling, grauman2022ego4d} and multiview cameras~\cite{kwon2021h2o, sener2022assembly101}. Such interactions are inherently fine-grained; recognizing them requires distinguishing subtle motions and object state changes. State-of-the-art methods for hand action recognition~\cite{feichtenhofer2019slowfast, patrick2021keeping, ma2022hand, wu2022memvit} primarily rely on multi- or single-view RGB streams, which are computationally heavy and unsuitable for resource-constrained scenarios like AR/VR. Motivated by advancements in lightweight hand pose estimation methods leveraging monochrome cameras~\cite{han2020megatrack, han2022umetrack, ohkawa2023assemblyhands}, and the evolution of low-dimensional sensors~\cite{liu2021neuropose, lee2024echowrist} such as accelerometers, MMG, EMG, demonstrating real-time hand pose estimation, we advocate the use of 3D hand poses as an input modality for recognizing hand-object interactions. Hand poses are a compact yet informative representation that captures the motions and nuances of hand movements.

\begin{figure*}[!t]
\centering 
\includegraphics[width=\textwidth, trim={1.25cm 9.5cm 7.75cm 6.25cm}, clip]{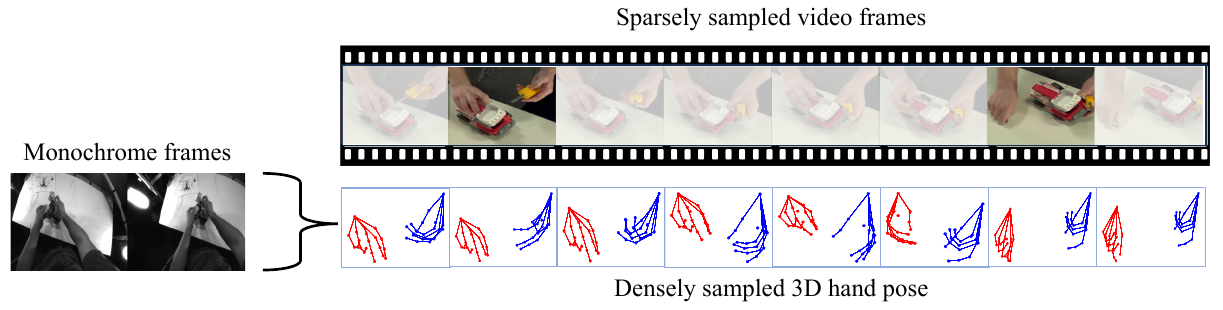}
\caption{
We densely sample 3D hand poses to understand fine-grained hand motions and sparsely sample RGB frames to capture the scene semantics. Our 3D hand poses are acquired from low-resolution monochrome cameras but can also come from  wearable sensors, facilitating an efficient understanding of hand-object interactions. Video frames and hand poses in the figure are from Assembly101~\cite{sener2022assembly101}.}
\label{fig:teaser}
\end{figure*}

\begin{figure*}[t]
 \centering
 \includegraphics[width=\textwidth, trim={1.5cm 8cm 1.5cm 0cm}, clip]{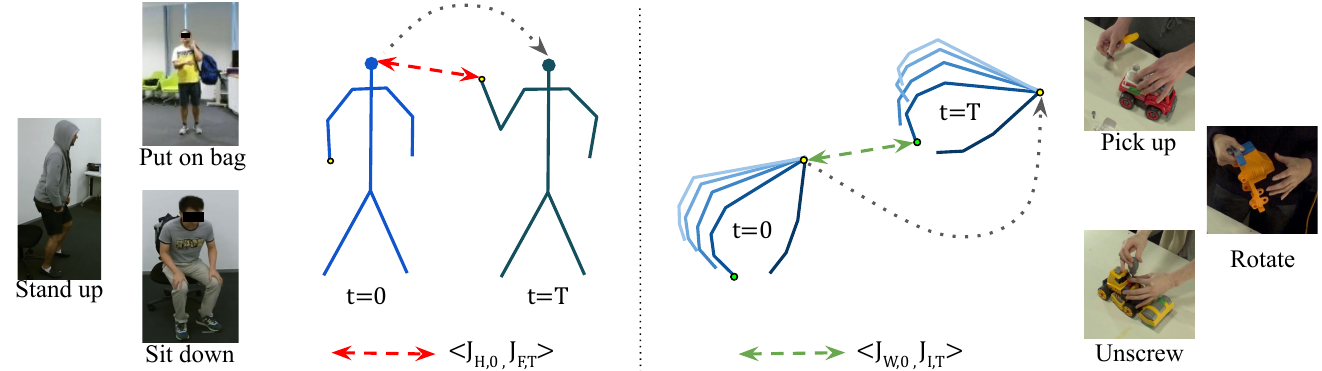}
 \caption{
 Comparing skeletal changes in full-body actions from NTU~RGB+D~120~\cite{liu2019ntu} (left) and hand actions from Assembly101~\cite{sener2022assembly101} (right). Two pose frames at interval $T$ are shown. $J_{j,t}$ indicates the 3D coordinate of joint $j$ at timestep $t$, and ${<}J_x,J_y{>}$ is the correlation between two such joints. Modeling the correlation between spatio-temporally distant joints can be informative for full-body poses but does not provide a useful action cue for hand poses.
 }
 \label{fig:hand_vs_body}
\end{figure*}

Existing works on 3D pose-based action recognition have focused primarily on full-body skeletons~\cite{yan2018spatial, liu2020disentangling, duan2022revisiting, wen2023interactive}.
Hand poses differ fundamentally from full-body skeletons. In full-body recognition datasets~\cite{shahroudy2016ntu, liu2019ntu}, the actions are predominantly static from a global perspective. The relative changes in joint or limb positions signify the action category, \eg, `sitting down'. Conversely, hand joints typically move together for many actions and lack a static joint as a global reference~\cite{sener2022assembly101}, \eg, `put down toy'. Full-body skeleton methods also benefit from modeling long-range spatiotemporal dependencies between joints~\cite{liu2020disentangling}, while this is less important for the hands as shown in~\cref{fig:hand_vs_body} and in \cref{sec:skel_vs_hand}. 

However, the 3D pose alone cannot encode the hands' actions. Unlike the full-body case, where actions are self-contained by the sequence of poses, the hands are often
manipulating objects~\cite{sener2022assembly101, kwon2021h2o, ohkawa2023assemblyhands}. 
Hand pose is excellent for identifying motions (verbs) but struggles with associated objects~\cite{sener2022assembly101}. Therefore, supplementing pose data with visual context from images or videos is crucial for full semantic understanding. However, as noted earlier, using dense RGB frames contradicts our 
motivation for using hand poses.

This paper introduces HandFormer, a novel and lightweight multimodal transformer that leverages dense 3D hand poses complemented with sparsely sampled RGB frames. To this end, we conceptualize an action as a sequence of short segments, termed \textit{micro-actions}, which are analogous to words that form sentence-level complex actions. Each micro-action comprises a dense sequence of pose frames and a single RGB frame.
As every hand joint moves in close spatial proximity, we encode the pose sequence from a Lagrangian view~\cite{rajasegaran2023benefits} and track each joint as an individual entity. 
HandFormer effectively models each micro-action 
whose features are then temporally aggregated for action classification.
This formulation benefits from our observations on the spatiotemporal dynamics of hand movement and delivers strong performance. Additionally, limiting long-range spatiotemporal dependencies among joints and only using sparse RGB frame features improves computational efficiency, making it well-suited for always-on, low-compute AR/VR applications.

Our contributions are: \textit{(i)} We analyze the differences between hand pose and full-body skeleton actions and design a novel pose sequence encoding that reflects hand-specific properties. 
\textit{(ii)} We propose HandFormer that takes a sequence of dense 3D hand pose and sparse RGB frames as input in the form of micro-actions. \textit{(iii)} HandFormer achieves state-of-the-art action recognition on H2O~\cite{kwon2021h2o} and Assembly101~\cite{sener2022assembly101}. Unimodal HandFormer with only hand poses outperforms existing skeleton-based methods in Assembly101, incurring at least $5\times$ fewer FLOPs. \textit{(iv)} We experimentally demonstrate how hand poses are crucial for multiview and egocentric action recognition.

\section{Related Work}
\label{sec:literature}

\textbf{Video Action recognition.} Video-based action recognition systems are well-developed with sophisticated 3D-CNN~\cite{tran2015learning, carreira2017quo, xie2018rethinking, feichtenhofer2019slowfast} or Video Transformer~\cite{arnab2021vivit, bertasius2021space, patrick2021keeping, liu2022video} architectures. However, they all bear significant computational expense for both feature extraction and motion modeling, either explicitly in the form of optical flow~\cite{simonyan2014two, carreira2017quo, kazakos2019epic} or implicitly through the architecture~\cite{arnab2021vivit, patrick2021keeping}. Such designs are well-suited for high-facility applications
but are not suitable for integration into lightweight systems. To this end, efficient video understanding is an active topic of research, with a focus on 
reducing expensive 3D operations~\cite{lin2019tsm, xie2018rethinking, tran2018closer, feichtenhofer2020x3d}, quadratic complexity of attention~\cite{jaegle2021perceiver, patrick2021keeping, liu2022video} and dropping tokens~\cite{fayyaz2022adaptive, bolya2022token}.
Additionally,~\emph{densly-sampled} RGB frames incur a high cost, yet limiting the temporal resolution will hinder fine-grained action understanding.
We propose complementing 3D hand poses with~\emph{sparsely-sampled} RGB frames, developing a lightweight video understanding system.

\noindent\textbf{Skeleton-based Action Recognition.}
Skeleton-based action recognition has been tackled with hand-crafted features \cite{wang2012mining, vemulapalli2014human}, sequence models like RNNs and LSTMs \cite{du2015hierarchical, zhang2017geometric}, and CNN-based methods that either employ temporal convolution on the pose sequence \cite{soo2017interpretable} or transform the skeleton data into pseudo-images to be processed with 2D or 3D convolutional networks \cite{hou2016skeleton, caetano2019skelemotion, duan2022revisiting}. Recent progress has been driven by GNN-based methods that exploit the graph structure of skeletal data to construct spatio-temporal graphs and perform graph convolutions~\cite{yan2018spatial, zhang2019graph}. These methods often model functional links between joints that go beyond skeletal connectivity~\cite{li2019actional, shi2019two} or expand the spatiotemporal receptive field~\cite{liu2020disentangling}. Self-attention and transformer-based methods have also been proposed~\cite{plizzari2021spatial, zhang2021stst, wen2023interactive}.
However, most existing methods are tailored for datasets involving full-body poses. Few works~\cite{hou2018spatial, li2021two, sabater2021domain, gan2023keyframe} study hand poses but are restricted to simple gestures, recognizable without complex temporal modeling.

\noindent\textbf{Fusing RGB with Skeleton.}
Skeletal data has been used for cropping images of body parts~\cite{cheron2015p}, for weighting RGB patches around regions of interest~\cite{bruce2021multimodal, bruce2022mmnet}, or for pooling CNN features~\cite{cao2017body, ahn2023star}. Projecting into common embedding space is done in~\cite{das2020vpn}, enabling pose distillation~\cite{das2021vpn++}. Multi-stream architectures are also designed with separate paths for RGB and skeleton having lateral connections between them~\cite{li2020sgm, weiyao2021fusion}. RGBPoseConv3D~\cite{duan2022revisiting} is a two-stream model that employs 3D~CNNs for both RGB and pose. Similar to skeleton-based action recognition, these multimodal approaches primarily focus on full-body poses. Some approaches simultaneously perform both hand pose estimation and action recognition, using pose data to supervise the training process~\cite{wen2023hierarchical, cho2023transformer}. However, it is important to note that pose estimation deals with lower-level semantics than action recognition. Actions can often be inferred even when the estimated poses are not accurate~\cite{ohkawa2023assemblyhands}.

\section{Modeling Full-body vs. Hand Skeletons}\label{sec:skel_vs_hand}
Existing skeleton-based action recognition datasets primarily consist of full-body poses where actions feature significant changes over time in limb positions relative to other body parts, which highly correlate to the action category. These changes can be captured through long-range spatiotemporal modeling using graph convolutions with large receptive fields~\cite{liu2020disentangling} or self-attention~\cite{wen2023interactive}. Conversely, hand actions show diverse movement patterns, as hands can move in arbitrary directions, often without significant changes in articulation. 
We analyze pose sequences from NTU RGB+D 120~\cite{liu2019ntu} and Assembly101~\cite{sener2022assembly101}, calculating the distance covered by each joint and identifying the least and the most active joints. We observe that, for full-body poses, these two representative joints show a significant difference in the distances covered, while for hand poses, the difference is subtle. The Pearson correlation coefficient is 0.93 for hand poses (indicating highly coupled joints) and 0.33 for full-body poses. Please refer to Suppl.~Sec.~A for details regarding these computations and comparisons.

An example showing the difference between body and hand poses is provided in~\cref{fig:hand_vs_body}, where we depict two frames separated by an interval of $T$ frames. For the full body skeleton, the head joint $J_H$ remains static, while the right hand joint $J_F$ moves upward. Conversely, for the hand, all the joints move, including the wrist joint $J_W$ and index fingertip $J_I$. The spatiotemporal correlation between the head joint in the first frame and the hand joint in the last frame (${<}J_{H,0}, J_{F,T}{>}$) is a crucial action cue that captures the relative structural change. However, such a correlation between the wrist and the fingertip for hand skeleton (${<}J_{W,0}, J_{I,T}{>}$) does not provide a stronger cue than the wrist movement itself. To avoid such redundant spatiotemporally distant correlations, we propose to divide the hand pose sequence into micro-action blocks of fixed temporal length. This formulation allows for encoding short-term movements while enabling parameter sharing.

Moreover, full-skeletal motion is a dominant feature in hand pose sequences, differentiating them from full-body poses. While the location and orientation of the human body can be trivial for most actions, this is not the case for hands. The hands do not conform to any particular 6D pose, and notably, they frequently and unpredictably alter their 6D poses over time. A comprehensive analysis of this phenomenon is provided in Suppl.~Sec.~A. To this end, we explicitly consider the global 6D poses of the hands during micro-action-based pose encoding.

\begin{figure*}[t!]
 \centering
 \includegraphics[width=\textwidth, trim={0cm 7cm 9cm 0cm}, clip]{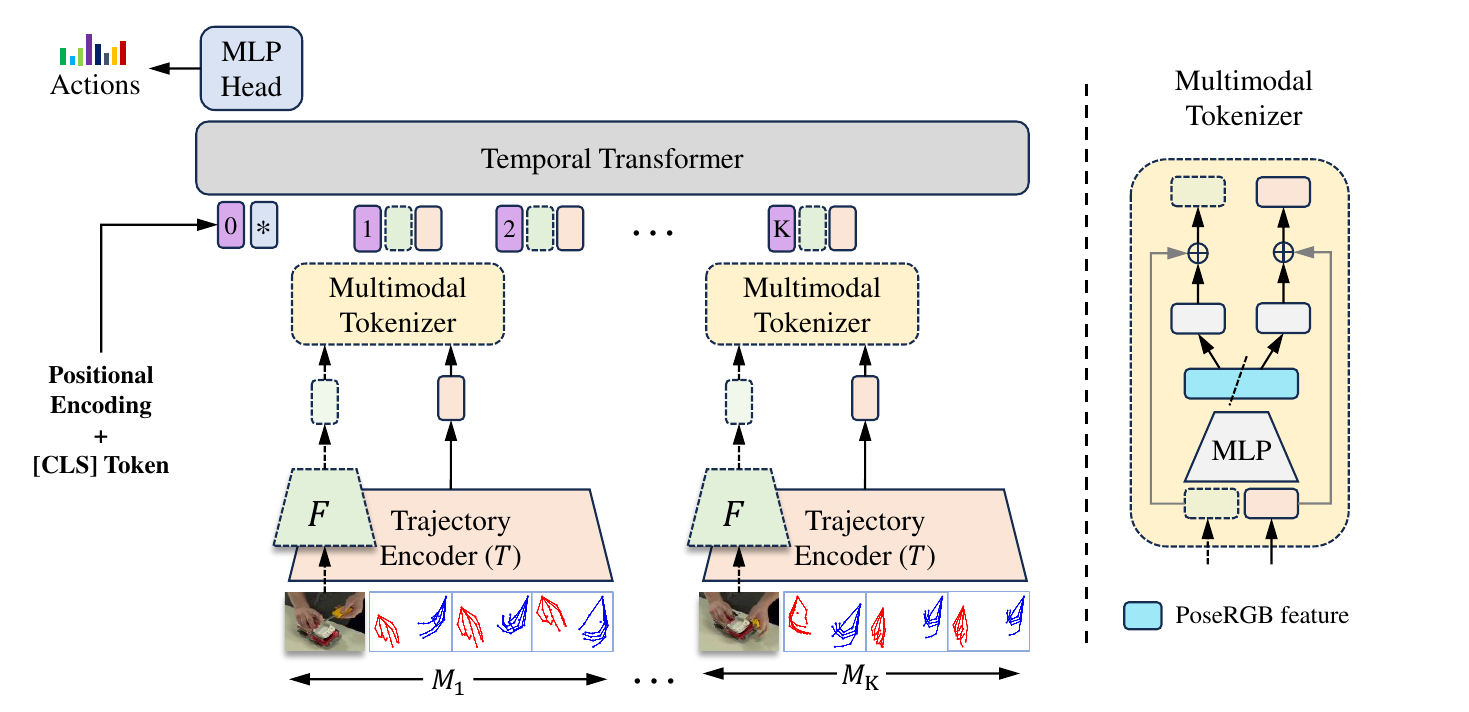}
 \caption{\textbf{Overall architecture of HandFormer.} An action segment is divided into $K$ micro-actions $\{M_1, M_2 \ldots M_K\}$. Each micro-action comprises a dense sequence of pose frames and a single RGB frame. The frame encoder $F$ and the trajectory encoder $T$ encode the RGB and the dense poses, respectively, after which they are passed to a Multimodal Tokenizer. The modality-mixed tokens are then fed to a Temporal Transformer.  
 Dotted paths are optional and only required when RGB is used.
 }
 \label{fig:pipeline}
\end{figure*}

\section{HandFormer}\label{sec:method}

\cref{fig:pipeline} illustrates our proposed HandFormer, which consists of a sequence of micro-action blocks, a novel trajectory encoder, and a temporal aggregation module. The design of our HandFormer allows us to easily incorporate semantic context by sampling a single RGB frame from certain micro-actions.

\subsection{Micro-actions}\label{ss:form_ma}

Given an action segment containing $\mathcal{T}$ frames sampled at a certain fps, the input to our model comprises --- \textit{i)} a dense sequence of 3D hand poses, represented as $\mathbf{S} = \{P_1, P_2, \ldots, P_\mathcal{T}\}$, where $P_t = \{P_t^{\text{\textit{left}}}, P_t^{\text{\textit{right}}}\}$ $\in \mathbb{R}^{2 \times J\times 3}$ signifying the 3D coordinates of $J$ keypoints in the left and right hands, respectively, and \textit{ii)} a sparse set of RGB frames sampled at intervals $\Delta f$, denoted as $\mathbf{V} = \{I_1, I_{1+\Delta f}, \ldots, I_{\mathcal{T}_{r}}\}$, where $I_t \in \mathbb{R}^{H\times W \times 3}$ are frame-wise RGB features and $\mathcal{T}_r = 1 + \left\lfloor\frac{\mathcal{T}-1}{\Delta f}\right\rfloor \times \Delta f$.

We factorize the raw input into a sequence of $K$ micro-action blocks of length $N$ frames, obtained by shifting the window across the action segment with a stride of $R$ frames. Each block consists of two components --- the initial appearance derived from the first RGB frame within the block and the hand motion characterized by the dense sequence of $N$ pose frames. To obtain a fixed length input containing $\mathcal{T}' = (K-1)\times R + N$ pose frames, we perform linear interpolation for each joint along the temporal axis, transforming pose sequence $\mathbf{S}$ having $\mathcal{T}$ frames to $\mathbf{S}^{\prime} = \{P^{\prime}_1, P^{\prime}_2, \ldots, P^{\prime}_{\mathcal{T}^{\prime}}\}$ having $\mathcal{T}^{\prime}$ frames, where $P^{\prime}_t \in \mathbb{R}^{2\times J\times 3}$. Thus, we represent the input as a sequence of micro-actions $\mathbf{M} = \{M_1, M_2, \ldots, M_{K}\}$, derived from $\mathbf{V}$ and $\mathbf{S}'$ through the following equations:
\begin{gather}
\begin{aligned}
 M_k &= \left[M_k^\text{RGB}, M_k^\text{Pose}\right] = \left[ I_{h(k)}, \{P^{\prime}_{g(k)+i}\}_{i=0}^{N-1}\right], 
\end{aligned}
\end{gather}
where $g(k) = (k-1)\times R + 1$ denotes the first pose frame in $k$-th Micro-action, and $h(k)$ determines the index of the nearest available RGB frame. 

The dense pose sequence in a micro-action captures fine-grained hand motion crucial for recognizing verbs, whereas a single RGB frame provides semantic context for recognizing objects. To extract features from micro-actions, we use a frame encoder $F$ and a trajectory encoder $T$, which operate on RGB frames and pose sequences, respectively. Consequently, the RGB and pose features for the $k^{th}$ micro-action is given by-
\begin{gather}
\begin{aligned}\label{eq:enc}
 \left[f_k^\text{RGB}, f_k^\text{Pose}\right] &= \left[F(M_k^\text{RGB}), T(M_k^\text{Pose})\right],
\end{aligned}
\end{gather}
where $f_k^\text{RGB}, f_k^\text{pose}\ \in \mathbb{R}^d$, where $d$ denotes the common dimensionality of both RGB and pose embedding space.

\subsection{Trajectory Encoder}\label{ss:traj_enc}

\begin{figure}[t]
 \centering
 \includegraphics[width=\textwidth, trim={0.5cm 9.5cm 1.5cm 2cm}, clip]{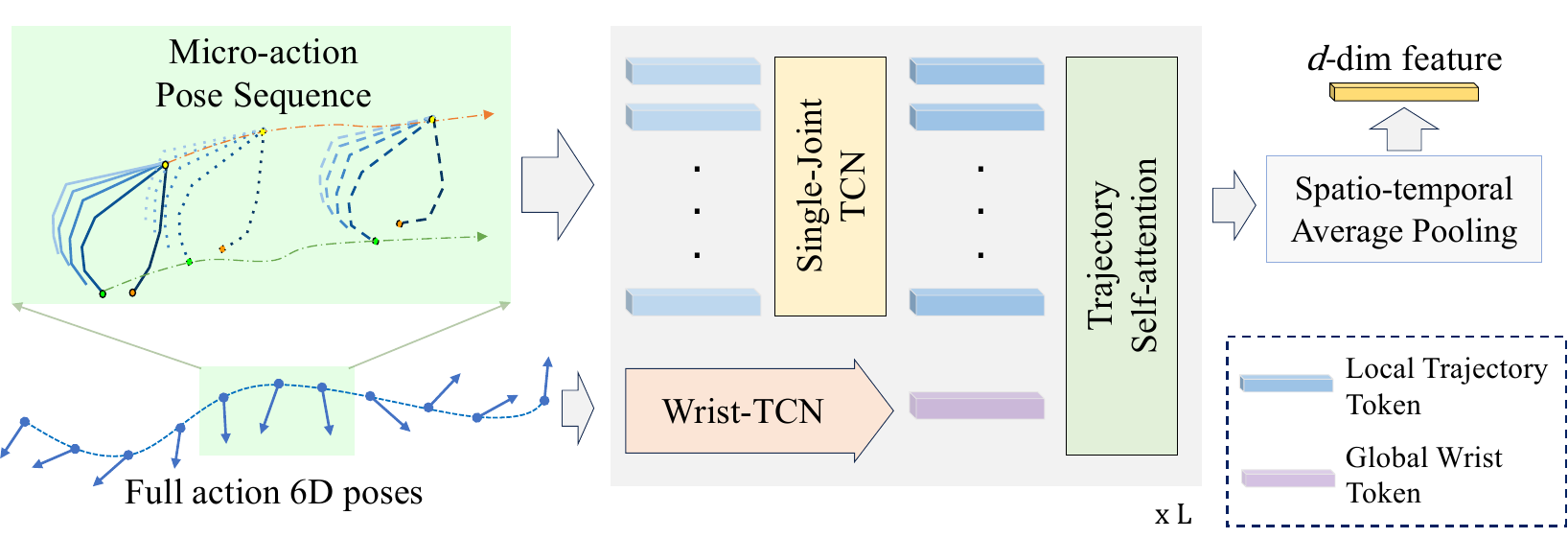}
 \caption{\textbf{Our Trajectory Encoder $T$}, which operates on micro-actions, derives tokens with trajectory-based features and performs self-attention to encode the pose sequence into a feature vector. The Single-Joint TCN is a Temporal Convolutional Network~\cite{lea2017temporal} that processes the trajectories of all the joints individually with shared parameters across all joints. Wrist-TCN takes an action-wide sequence of wrist location and hand orientation (6D pose) to produce a global reference token.}
 \label{fig:mha_tokens}
\end{figure}

We devise a trajectory-based pose encoder to encode dense hand pose sequences within micro-actions, illustrated in~\cref{fig:mha_tokens}. Each joint is represented by its trajectory of dimension $3\times N$, encapsulating the sequence of 3D coordinates across the $N$ pose frames of a micro-action. This yields $2\times J$ feature vectors for the $J$ joints of two hands. Each joint's trajectory is passed through a TCN~\cite{lea2017temporal}, whose parameters are shared for all joints. This produces $2\times J$ Local Trajectory Tokens. Additionally, the full-skeletal motion of the hand during the entire action is used as a reference through an additional token named Global Wrist Token. This token is generated by a separate TCN operating on the sequence of 6D poses of hands, indicating wrist location and hand orientation. Subsequently, a self-attention layer is applied to these trajectory tokens, preserving the temporal dimension for subsequent stages. This iterative process culminates in spatiotemporal average pooling, summarizing the hand motion of the micro-action.

\subsection{Multimodal Tokenizer}

This section discusses incorporating sparsely sampled RGB frames into HandFormer to better capture scene semantics. A single frame is sampled from each micro-action, generating an extended crop ($1.25\times$) around the hands. While the full image provides an overall scene context, the crop focuses specifically on hand-object interaction regions~\cite{chatterjee2023opening}. Features for both are separately generated using a pre-existing image encoder and then aggregated to enrich the hand-object interaction feature with scene context. The hand-object ROI crop can be obtained using an off-the-shelf HOI detector \cite{shan2020understanding} or by using the corresponding hand pose projections. We opt for the latter.

Our multimodal tokenizer in~\cref{fig:pipeline} generates RGB and Pose tokens enhanced via multimodal interactions. This involves concatenating each frame feature and trajectory encoding, and projecting them into a shared PoseRGB feature space using an MLP. This PoseRGB feature is then split into two parts, which are added to the original frame feature and trajectory encoding, respectively.

\subsection{Temporal Transformer}

The multimodal tokenizer provides RGB and pose tokens for each micro-action. Since an action segment consists of a sequence of micro-actions, these micro-action tokens are aggregated over time via a temporal transformer. A video $\mathbf{V}$, divided into $K$ micro-actions can be represented by two sets of tokens $\{\hat{f}_k^\text{RGB}\}_{k=1}^K$ and $\{\hat{f}_k^\text{Pose}\}_{k=1}^K$, respectively, produced by the multimodal tokenizer. These $2\times K$ tokens of dimension $d$ form the input sequence of the temporal transformer. Positional encoding and modality embedding are added to each input token to indicate the temporal location and source modality. We use the fixed sine/cosine positional encoding~\cite{vaswani2017attention}, assigning the same position to two tokens from the same micro-action. Modality embeddings are learned and shared across the tokens from the same modality (RGB or Pose). Following standard practice~\cite{devlin2018bert, dosovitskiy2020image}, we prepend an additional learnable class token [CLS] $\in \mathbb{R}^d$, the output of which is then used for classifying actions.

\subsection{Learning Objectives}
HandFormer is trained end-to-end for action recognition, supervised by a cross-entropy loss: $\mathcal{L}_{cls} = - a_i\log \hat{a}_i$, where $a_i$ represents the ground truth action label for the $i^{th}$ sample, and $\hat{a}_i$ denotes the predicted action category. The input pose and RGB modalities provide complementary information regarding a scene, capturing motion and interacting objects, respectively. To effectively utilize this, we employ explicit verb and object supervision~($\mathcal{L}_{verb}$, $\mathcal{L}_{obj}$) via two additional learnable class tokens. 
Given that pose strongly correlates with verbs, whereas objects are identifiable from RGB frames alone, we let the verb class token attend exclusively to pose encodings and the object class token to frame features.

\noindent\textbf{Feature Anticipation Loss.}
Hand pose sequence captures the primary sources of state changes during hand-object interactions.
We posit that the visual state from an initial RGB frame, in combination with the subsequent hand pose sequence, is indicative of the visual state that results from the completion of the sequence. 
Therefore, given an RGB feature and the corresponding pose features from a micro-action, we force our model to anticipate the RGB feature for the next micro-action by minimizing an $L_1$ feature loss. This loss, inspired by existing efforts~\cite{vondrick2016anticipating, girdhar2021anticipative}, quantifies the difference between the \emph{anticipated} image feature and the true feature extracted from a frozen image encoder. Formally, 

\begin{gather}
\begin{aligned}\label{eq:rec_loss}
\mathcal{L}_{ant} = \sum_{k=1}^{K-1}\big\lVert \Phi_{\text{ant}}(f_k^{\text {PoseRGB}})-f_{k+1}^{\text{RGB}} \big\rVert_1
\end{aligned}
\end{gather}
where $ \Phi_{\text{ant}}$ denotes a linear projection layer. Hence, the total loss is-

\begin{equation}\label{eq:av_loss}
\mathcal{L} = \mathcal{L}_{cls} + \lambda_1 \mathcal{L}_{verb} + \lambda_2 \mathcal{L}_{obj} + \lambda_3 \mathcal{L}_{ant}
\end{equation}
where $\lambda_1, \lambda_2, \lambda_3$ are hyperparameters used to balance the four losses.

\section{Experiments}
\label{sec:experiments}
\textbf{Datasets.}
We conduct experiments on two publicly available hand-object interaction datasets. 
\textbf{Assembly101} \cite{sener2022assembly101} is a large-scale multiview dataset that features videos of procedural activities for assembling and disassembling 101 take-apart toy vehicles. This dataset has 1380 fine-grained actions based on 24 verbs and 90 objects. It consists of 12 temporally synchronized views --- eight static and four egocentric. The egocentric views exhibit low-resolution monochrome images, making it challenging even for human eyes to discern objects in view. Hence, we opt for fixed views as our RGB modality.
3D hand poses estimated using off-the-shelf MEgATrack~\cite{han2020megatrack} are provided by the dataset. 
\textbf{H2O}~\cite{kwon2021h2o} features four participants performing 36 actions that involve 8 objects and 11 verbs.
Ground truth 3D hand poses 
are provided. 
The dataset consists of 5 temporally synchronized RGB views --- 4 fixed and 1 egocentric. We use RGB frames from the egocentric view only.

\noindent\textbf{Implementation Details.} Similar to~\cite{wen2023interactive}, our pose input consists of 120 frames by setting $\mathcal{T}'=120$. The number of frames per micro-action ($N$) is 15. For multimodal experiments, we set the window stride $R=N$ and take $K=8$ non-overlapping micro-actions, which allows us to sample 8 RGB frames for each video following~\cite{sener2022assembly101}. However, we allow a 50\% overlap between consecutive micro-actions for pose-only variants. 
A frozen ViT~\cite{dosovitskiy2020image} or ResNet~\cite{he2016deep}, followed by a learnable linear layer, is used as the frame encoder $F$. More details are provided in Suppl.~Sec.~C.
Each model is trained for 50 epochs using SGD with momentum 0.9, a batch size of 32, a learning rate of 0.025, and step LR decay with a factor of 0.1 at epochs \{25, 40\}.
Loss hyperparameters $\{\lambda_1, \lambda_2, \lambda_3\}$ are chosen to be $\{1.0,1.0,2.0\}$.

\noindent\textbf{Model Variants.} We propose several variants of our model by adjusting the width $d$ and the number of layers $T_n$ of the transformer to balance efficiency and accuracy. Our default HandFormer, denoted \mbox{HandFormer-B}, has parameters $(d, T_n)=(256,2)$. We introduce a larger variant, HandFormer-L, with $(d, T_n)=(512,4)$. We also explore different configurations for the number of input joints $J$ per hand. Unless otherwise mentioned, we utilize all 21 joints per hand along with the base model denoted as HandFormer-B/21 while offering a highly efficient option utilizing only six joints per hand (five fingertips and the wrist) termed HandFormer-B/6. Additionally, HandFormer-X/J$\times\mathcal{N}$ denotes a multimodal variant that utilizes $\mathcal{N}$ RGB frames, where $\mathcal{N} \le K$.

\subsection{Comparison with State-of-the-Art}
To evaluate the effectiveness of our proposed architecture, we compare it against several baselines. For pose-only comparisons, we choose a graph-based network MS-G3D~\cite{liu2020disentangling} and an attention-based method ISTA-Net~\cite{wen2023interactive} --- the two best performing skeleton-based methods on Assembly101~\cite{sener2022assembly101}, as reported by~\cite{wen2023interactive}. We also employ video baselines TSM \cite{lin2019tsm} and SlowFast~\cite{feichtenhofer2019slowfast}, emphasizing efficiency and high temporal resolution, respectively. Additionally, we include H2OTR~\cite{cho2023transformer}, the state-of-the-art for the H2O dataset. For Assembly101, we replicate the results of the methods above using RGB frames from \textit{view 4} of the dataset, while results on H2O are acquired from the respective papers. Furthermore, on both datasets, we train and test RGBPoseConv3D~\cite{duan2022revisiting}, which is current state-of-the-art in multimodal action recognition with skeleton and RGB data. 

We evaluate three variants of our method by controlling the input modalities. As shown in \cref{tab:sota}, our unimodal pose-only model excels in verb recognition on Assembly101. In contrast, RGB-based methods benefit from object appearance and usually perform well for actions due to the strong object recognition.

In this context, the accuracy of ISTA-Net on H2O is not directly comparable to our method, as they also use the 6D object poses, while we only use hand poses as input. RGBPoseConv3D struggles to achieve satisfactory performance, particularly on Assembly101, suggesting that generalizing to hand poses is non-trivial for skeleton-based methods. Therefore, we combine two best-performing unimodal methods with late fusion (MS-G3D + TSM) to create a stronger baseline. Our model even outperforms this, indicating the effectiveness of our proposed multimodal fusion.

\begin{table}[t]
 \centering
 \caption{\textbf{Quantitative comparison with state-of-the-art methods} on Assembly101~\cite{sener2022assembly101} and H2O~\cite{kwon2021h2o}. For H2O, 6D object pose is used by ISTA-Net during training and inference and by H2OTR during training. `MS-G3D + TSM' denotes a late fusion of the corresponding unimodal architectures. Reported GFLOPs include pose estimation costs in the context of Assembly101 at 15 fps. \textsuperscript{\textdagger}For pose-only methods, pose estimation costs are common and excluded for simpler comparison.}
 \begin{adjustbox}{width=0.89\linewidth}
 \begin{tabular}{@{}llccccccc@{}}
 \toprule
 \multicolumn{2}{l}{\multirow{2}{*}{\textbf{Method}}} & \multirow{2}{*}{\textbf{Pose}} & \multirow{2}{*}{\quad\textbf{RGB}\quad} & \multirow{2}{*}{\quad\textbf{GFLOPs}\quad} & \multicolumn{3}{c}{\textbf{Assembly101}} & \quad\textbf{H2O} \\ \cline{6-8}
 & & & & & \textbf{Action} & \textbf{Verb} & \textbf{Object} & \quad\textbf{Action}\\
 \midrule
 \multicolumn{2}{l}{MS-G3D \cite{liu2020disentangling}} & \cmark & \xmark & 21.2\textsuperscript{\textdagger} & 28.78 & 63.46 & 37.26 & \quad50.83 \\
 \multicolumn{2}{l}{ISTA-Net \cite{wen2023interactive}} & \cmark & \xmark & 35.2\textsuperscript{\textdagger} & 28.14 & 62.70 & 36.77 & \quad89.09 \\
 \multicolumn{2}{l}{SlowFast \cite{feichtenhofer2019slowfast}} & \xmark & \cmark & 65.7 & - & - & - & \quad77.69 \\
 \multicolumn{2}{l}{TSM \cite{lin2019tsm}} & \xmark & \cmark & 33.0 & 35.27 & 58.27 & 47.45 & \quad- \\
 \multicolumn{2}{l}{H2OTR\cite{cho2023transformer}} & \xmark & \cmark & - & - & - & - & \quad90.90 \\ 
 \multicolumn{2}{l}{RGBPoseConv3D \cite{duan2022revisiting}\quad\quad} & \cmark & \cmark & 68.9 & 33.61 & 61.99 & 42.90 & \quad83.47 \\
 \multicolumn{2}{l}{MS-G3D + TSM} & \cmark & \cmark & 66.2 & 39.74 & 65.12 & 51.12 & \quad- \\
 \midrule
 \multirow{3}{*}{\rotatebox[origin=c]{0}{\textbf{HandFormer-B/21$\times$8}\quad}} & & \cmark & \xmark & 4.2\textsuperscript{\textdagger} & 28.80 & 65.33 & 36.28 & \quad57.44\\
 & & \xmark & \cmark & 33.0 & 32.07 & 55.61 & 44.89 & \quad84.71\\
 & & \cmark & \cmark & 47.6 & \textbf{41.06} & \textbf{69.23} & \textbf{51.17} & \quad\textbf{93.39} \\
 \bottomrule
 \end{tabular}
 \end{adjustbox}
 \label{tab:sota} 
\end{table}

\subsection{Skeleton-based Action Recognition for Hands} 
The compositional nature of action classes in hand-object interaction videos allows us to break down the action into a verb and an object. 
Recognizing such actions from 3D hand poses is an ill-posed problem, as the hand skeletons lack explicit information about the interacting objects, which are also part of the action semantics. 
However, as the pose data completely captures the hand motion information, it can be reliably used for verb recognition. Therefore, we evaluate a pose-only version of our method on the verb recognition task and compare the performance and efficiency metrics with other state-of-the-art skeleton-based methods in~\cref{fig:pose_verb}. Our method uses significantly fewer GFLOPs due to our spatiotemporal factorization using micro-actions.
With $J=21$, all our HandFormer variants outperform existing methods. The $J=6$ variant is exceptionally efficient with comparable accuracy, notably surpassing the efficiency-focused Shift-GCN~\cite{cheng2020skeleton}.
HandFormer-H/21 combines the HandFormer-B and HandFormer-L variants and becomes our best-performing model, improving over MS-G3D by 2.04\% while maintaining comparable FLOPs.

\subsection{How many RGB frames are required?} 
In our model, the pose modality maintains a high temporal resolution to capture fine-grained hand movements, resulting in good verb recognition performance. 
On the contrary, RGB frames are primarily required to introduce semantic context for object recognition and does not necessitate a high temporal resolution like hand movements. 
The design of our model allows us to sample only a few RGB frames (as low as one) but still perform competitively at a reduced computational cost.
~\cref{fig:acc_rgb_variation} shows the impact of using more RGB frames for Assembly101~\cite{sener2022assembly101}. For this experiment, non-overlapping micro-actions are considered. Using only one RGB frame in HandFormer (35.46) outperforms the video model TSM (35.27), as shown in~\cref{tab:sota}. This performance gain stems primarily from a significant improvement in object accuracy, with only a slight enhancement in verb accuracy. However, including more RGB frames shows a diminishing return, which is unsurprising as additional frames are expected to provide redundant information. These results are obtained with a simplified version of our model by setting $\mathcal{L} = \mathcal{L}_{cls}$ and bypassing the multimodal tokenizer.

\begin{table}[t]
\begin{minipage}[t]{.46\textwidth}
\centering

 \includegraphics[width=\linewidth]{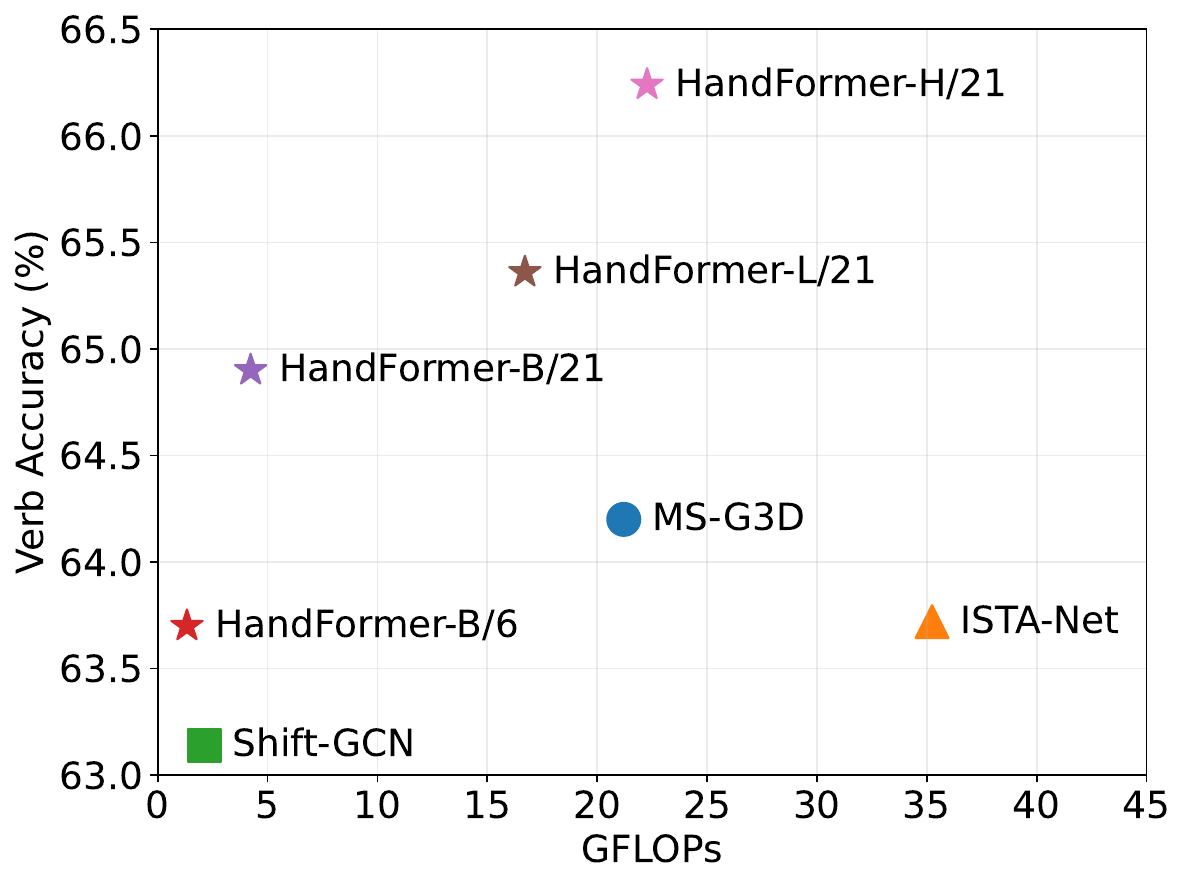}
 \captionof{figure}{\textbf{Comparison of skeleton-based methods for verb recognition} on Assembly101~\cite{sener2022assembly101}. Our method achieves state-of-the-art performance while utilizing significantly fewer FLOPs.}\label{fig:pose_verb}

\end{minipage}\hfill
\begin{minipage}[t]{.44\textwidth}
 \centering
 \includegraphics[width=\linewidth]{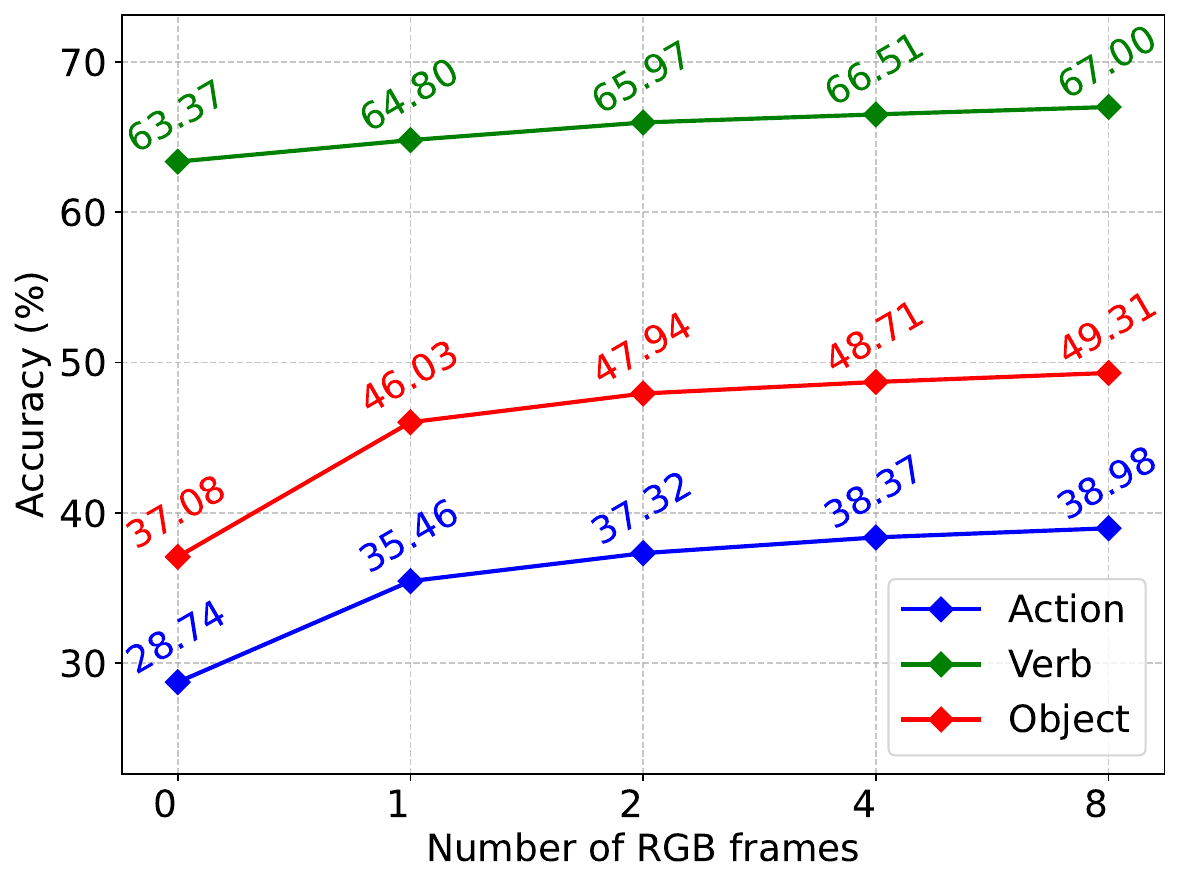}
 \captionof{figure}{\textbf{Ablating number of RGB frames in HandFormer} on Assembly101~\cite{sener2022assembly101}. With more RGB frames, verb recognition shows marginal gain, whereas object recognition shows improvement with diminishing returns.}\label{fig:acc_rgb_variation}
 \end{minipage} 
\end{table}

\subsection{Can 3D hand pose be an efficient alternative to multi-view?}\label{ss:multiview}
While multiview action recognition benefits from precise hand-movement information in 3D space, processing all views with video models is expensive and highly redundant. In \cref{tab:3d_vs_multi}, we demonstrate that combining 3D hand pose with a single RGB view (\textit{view 4}) achieves comparable performance to multiview action recognition on Assembly101~\cite{sener2022assembly101}. Specifically, our action recognition performance (`Single View + \textit{Pose}') matches the fusion of the two most informative views --- \textit{view 1} and \textit{view 4}. 
Notably, our verb recognition accuracy is on par with combining all 8 RGB views, though the latter incurs at least $5\times$ more FLOPs despite using the efficient TSM model. 
Additionally, using hand pose in combination with an RGB view (`Single View + \textit{Pose}') outperforms directly using the egocentric videos (`Single View + \textit{Egocentric}') from which the hand poses are derived. While fusing multiple RGB views improves accuracy by ensembling multiple complementary predictions, the computational overhead increases significantly. In contrast, our model processes hand pose and single-view RGB frames, enhancing efficiency by leveraging the less redundant and low-dimensional pose data. 
 
\noindent
\begin{table}[t]
\begin{minipage}[t]{0.99\linewidth}
 \begin{minipage}[t]{0.45\linewidth}
 \centering
 \captionof{table}{\textbf{Multi-view Action Recognition} on Assembly101\cite{sener2022assembly101}. HandFormer with `Single View+\textit{Pose}' has better performance than `Single View+\textit{Egocentric}'. Also, verb performance is comparable to using all (8) RGB views.}\label{tab:3d_vs_multi}
 \resizebox{\linewidth}{!}{
 \begin{tabular}{lccc@{}}
 \toprule
 \textbf{Views }& \textbf{Action} & \textbf{Verb} & \textbf{Object} \\
 \midrule
 Single View& 35.27 & 58.27 & 47.45 \\
 Single View + \textit{Egocentric} & 37.75 & 61.80 & 49.43 \\
 Two Views & 41.96 & 65.22 & 53.26 \\
 All (8) Views & 47.51 & 70.99 & 57.73 \\
 \midrule
 Single View + \textit{Pose} & 41.06 & 69.23 & 51.17 \\
 \bottomrule
 \end{tabular}}
 \end{minipage}
 \hfill
 \begin{minipage}[t]{0.52\linewidth}
 \centering
 \captionof{table}{\textbf{Cross-view performance of HandFormer} in Assembly101~\cite{sener2022assembly101} shows its generalization capability to unseen \textit{view 1}, outperforming the video baseline which is trained on \textit{view 1} directly. Egocentric views are the source of hand poses here and, therefore, are included in the video models.}\label{tab:cross_view}
 \resizebox{\linewidth}{!}{
 \begin{tabular}{@{}lccccc@{}}
 \toprule
 \textbf{Method} & \textbf{Train on} & \textbf{Test on} & \textbf{Action} & \textbf{Verb} & \textbf{Object}\\
 \midrule
 \multirow{3}{*}{TSM \cite{lin2019tsm}} & \textit{v4} + ego & \textit{v4} + ego & 37.75 & 61.80 & 49.43 \\
 & \textit{v4} + ego & \textit{v1} + ego & 35.27 & 59.53 & 47.72 \\
 & \textit{v1} + ego & \textit{v1} + ego & 36.21 & 60.78 & 48.52 \\ 
 \midrule
 \multirow{2}{*}{Our Method} & \textit{v4} + Pose \quad & \textit{v4} + Pose & 41.06 & 69.23 & 51.17 \\
 & \textit{v4} + Pose \quad & \textit{v1} + Pose & 38.43 & 67.86 & 48.32 \\
 \bottomrule
 \end{tabular}}
 \end{minipage}
 \end{minipage} 
\end{table}

\noindent\textbf{Cross-view Generalization.} 3D hand pose offers a unique opportunity for cross-view generalization because of its universality across different viewpoints. To evaluate the effectiveness of our method for unseen views, we train our model with frame-wise RGB features from \textit{view 4} and test it on \textit{view 1}. We train TSM~\cite{lin2019tsm} as a baseline video model on both RGB views separately. As our method includes 3D hand poses obtained from the egocentric views, we include the egocentric videos in the TSM baselines for a fair comparison. As seen from~\cref{tab:cross_view}, our method, trained on \textit{view 4}, generalizes well on unseen \textit{view 1}, outperforming the TSM model that was directly trained on \textit{view 1}.

\subsection{Egocentric Action Recognition}\label{ss:ego}
Recognizing actions in egocentric videos is challenging due to camera motion and the occlusion of interacting objects by the hands. Additionally, in the case of Assembly101~\cite{sener2022assembly101}, the egocentric cameras are similar to Oculus Quest VR headsets, which provide monochrome low-resolution frames. As a result, action recognition performance is significantly lower than fixed RGB views, as found in \cite{sener2022assembly101}. We address this challenging scenario with our proposed multimodal architecture and achieve state-of-the-art performance in egocentric action recognition on Assembly101. For this experiment, we used frame-wise TSM features provided by~\cite{sener2022assembly101}. As depicted in \cref{tab:ego_view}, our model, using a single egocentric view (\textit{e3} or \textit{e4}) outperforms the fusion of four egocentric views as reported in \cite{sener2022assembly101}. Moreover, fusing \textit{e3} and \textit{e4} significantly enhances our model, resulting in a 4.25\% increase in action accuracy over the baseline.

\subsection{Ablation Studies}
\noindent\textbf{Keypoints.} Not all joints of the hands are equally informative for understanding hand actions. 
For instance, fingertips exhibit greater mobility compared to the inner joints. Moreover, from an egocentric viewpoint, certain joints are more prone to self-occlusion than others. In~\cref{tab::kp_ablate}, we present the impact of incorporating varying numbers of joints on verb recognition within the Assembly101 dataset~\cite{sener2022assembly101}. For the case of 6 joints, we consider only the wrist joint along with the five fingertips. To expand to 11 joints, we incorporate all the joints along the index and the thumb, which are least affected by self-occlusion. We also show the effect of including the Global Wrist Token, which acts as a reference to the global motion of the hands while encoding micro-actions.

\noindent
\begin{figure}[t]
\begin{minipage}[t]{0.98\linewidth}
 \begin{minipage}[t]{0.55\linewidth}
 \centering
 \captionof{table}{\textbf{Egocentric action recognition} in Assembly101 \cite{sener2022assembly101}. TSM features from \cite{sener2022assembly101} are used as RGB frame features.}\label{tab:ego_view}
 \resizebox{\linewidth}{!}{
 \begin{tabular}{@{}lccc@{}}
 \toprule
 \textbf{Method} & \textbf{Action} & \textbf{Verb} & \textbf{Object}\\
 \midrule
 TSM egocentric (fuse 4 views) $\quad$ & 33.80 & 59.00 & 46.50 \\ 
 \midrule
 Egocentric (\textit{e3}) + Pose & 36.07 & 65.52 & 45.82 \\
 Egocentric (\textit{e4}) + Pose & 35.56 & 65.79 & 45.20 \\
 Egocentric (\textit{e3$+$e4}) + Pose & \textbf{38.05} & \textbf{66.32} & \textbf{47.86} \\
 \bottomrule
 \end{tabular}}
 \end{minipage}
 \hfill
 \begin{minipage}[t]{0.40\linewidth}
 \centering
 \captionof{table}{\textbf{Keypoint ablation} for verb recognition in Assembly101~\cite{sener2022assembly101}}\label{tab::kp_ablate}
 \resizebox{0.95\linewidth}{!}{
 \begin{tabular}{@{}ccc@{}}
 \toprule
 \multirow{2}{*}{\textbf{~\#Joints~}} & \textbf{Global} & \textbf{Verb}\\
 & \textbf{~Reference~} & \textbf{Accuracy (\%)}\\
 \midrule
 21 & \xmark & 64.17\\
 21 & \cmark & \textbf{64.90}\\
 11 & \cmark & 64.77\\
 6 & \cmark & 63.70\\
 \bottomrule
 \end{tabular}}
 \end{minipage}
\end{minipage}
\end{figure}

\noindent\textbf{Micro-action length.} As the resized input comprises a fixed number of pose frames, enlarging the window size for a micro-action decreases the number of micro-actions to aggregate, and vice-versa. In~\cref{tab::ablate_wsize}, we vary the micro-action length for verb recognition in the Assembly101~\cite{sener2022assembly101} using 6 joints per hand as input, \ie fingertips and wrist joint. The input pose clip is temporally resized to $\mathcal{T}'=120$ before breaking into micro-actions.
Lengths 1 and 120 represent two extreme versions with frame-based and trajectory-based encoding, respectively, while the others conform to our micro-action-based formulation.

\noindent
\begin{figure}[t]
\begin{minipage}[t]{0.99\linewidth}
 \begin{minipage}[t]{0.52\linewidth}
 \centering
 \captionof{table}{\textbf{Micro-action length ablation} for verb recognition in Assembly101~\cite{sener2022assembly101}}\label{tab::ablate_wsize}
 \resizebox{\linewidth}{!}{
 \begin{tabular}{@{}lrrrrrr@{}}
 \toprule
 \textbf{\#Frames} & \textbf{1} & \textbf{15} & \textbf{30} & \textbf{60} & \textbf{120} \\
 \midrule
 \textbf{Verb} & \mr{2}{\quad59.12} & \mr{2}{\quad\textbf{63.70}} & \mr{2}{\quad63.68} & \mr{2}{\quad63.51} & \mr{2}{\quad62.29} \\
 \textbf{Accuracy(\%)} &&&&&\\
 \bottomrule
 \end{tabular}}
 \end{minipage}
 \hfill
 \begin{minipage}[t]{0.43\linewidth}
 \centering
 \captionof{table}{\textbf{Ablating unimodal temporal aggregation} for verb recognition in Assembly101~\cite{sener2022assembly101}
 }\label{tab::ablate_tempAgg}
 \resizebox{0.95\linewidth}{!}{
 \begin{tabular}{@{}lrrr@{}}
 \toprule
 \textbf{Temp. Agg.} & \textbf{TCN} & \quad \textbf{LSTM} & \quad \textbf{Transformer} \\
 \midrule
 \textbf{Verb Acc. (\%)} & \quad62.95 & 63.34 \quad& \textbf{63.70}\quad\quad\quad \\
 \bottomrule
 \end{tabular}}
 \end{minipage}
\end{minipage}
\end{figure}

\noindent\textbf{Temporal Aggregation.} After extracting micro-action features, aggregation for action recognition can be done with any sequence model. In~\cref{tab::ablate_tempAgg}, we evaluate the effectiveness of different temporal aggregation methods for verb recognition in the Assembly101 dataset~\cite{sener2022assembly101}. Here, we use 6 joints per hand as input, \ie, the fingertips and the wrist joint.

\noindent\textbf{Loss components.} To assess the individual contributions of different components, we begin with a basic configuration. We then systematically introduce each element to understand its impact on the overall performance as observed in \cref{tab:ablate}. Incorporating modality interaction between RGB and pose at the micro-action level through a multimodal tokenizer enhances action accuracy. The introduction of auxiliary losses also has a positive impact, resulting in an overall improvement of 2.08\% for Assembly101 \cite{sener2022assembly101} and 7.44\% for H2O \cite{kwon2021h2o}.

\begin{table}[t]
 \centering
 \caption{Ablating tokenization and different losses for action recognition}
 \label{tab:ablate}
 \begin{adjustbox}{width=0.9\linewidth}
 \begin{tabular}{@{}ccccc@{}}
 \toprule
 \textbf{\quad Multimodal\quad} & \textbf{\quad Feature\quad} & \textbf{\quad Verb \& Object\quad} & \multicolumn{2}{c}{\textbf{Action Accuracy (\%)}} \\ \cline{4-5}
 \textbf{Tokenizer} & \textbf{Ant. Loss} & \textbf{Loss} & \textbf{Assembly101}~\cite{sener2022assembly101} & \textbf{H2O}~\cite{kwon2021h2o} \\
 \midrule
 \xmark & \xmark & \xmark & 38.98 & 85.95\\
 \cmark & \xmark & \xmark & 40.19 & 88.84\\
 \cmark & \cmark & \xmark & 40.24 & 89.26 \\
 \xmark & \xmark & \cmark & 40.56 & 90.50\\
 \cmark & \cmark & \cmark & \textbf{41.06} & \textbf{93.39} \\
 \bottomrule
 \end{tabular}
 \end{adjustbox}
\end{table}

\section{Conclusion}\label{sec:conclusion}
With the growing interest in AR/VR and wearables, hand pose estimation has rapidly advanced,
and holds promise as a compact and domain-independent modality to complement the visual input. To address the underexplored domain of using 3D hand pose as a modality for hand-object interaction recognition, 
we introduce HandFormer, a novel multimodal transformer that leverages dense sequences of 3D hand poses with sparsely sampled RGB frames to achieve state-of-the-art action recognition performance. Our model also reduces computational requirements, offering immediate significance across various low-resource applications in mobile devices.

\noindent \textbf{Limitations.} Our method relies on the availability of hand poses, which, if extracted from the visual modality with pose estimation tools~\cite{han2020megatrack,han2022umetrack}, can encounter out-of-view scenarios and produce noisy poses. Our experiments reveal that these estimated poses can still achieve good accuracy, yet further research can be conducted to explicitly address this phenomenon. We also assume that a uniform sampling of RGB frames from each micro-action should provide good representations for understanding the semantic context. However, not all frames are equally important in understanding the action. In such a case, adaptive frame sampling methods can be employed, which we leave for future work.

\section*{Acknowledgements}
This research is supported by A*STAR under its National Robotics Programme (NRP) (Award M23NBK0053).

\renewcommand{\thesection}{\Alph{section}}
\renewcommand{\theHsection}{\thesection}
\section*{Appendix}
\setcounter{section}{0}

\section{Statistical Analysis: Full-body vs. Hand Skeletons}

\begin{figure*}[!b]
    \centering
    \includegraphics[width=0.8\textwidth, trim={1cm 3.5cm 8cm 3cm}, clip]{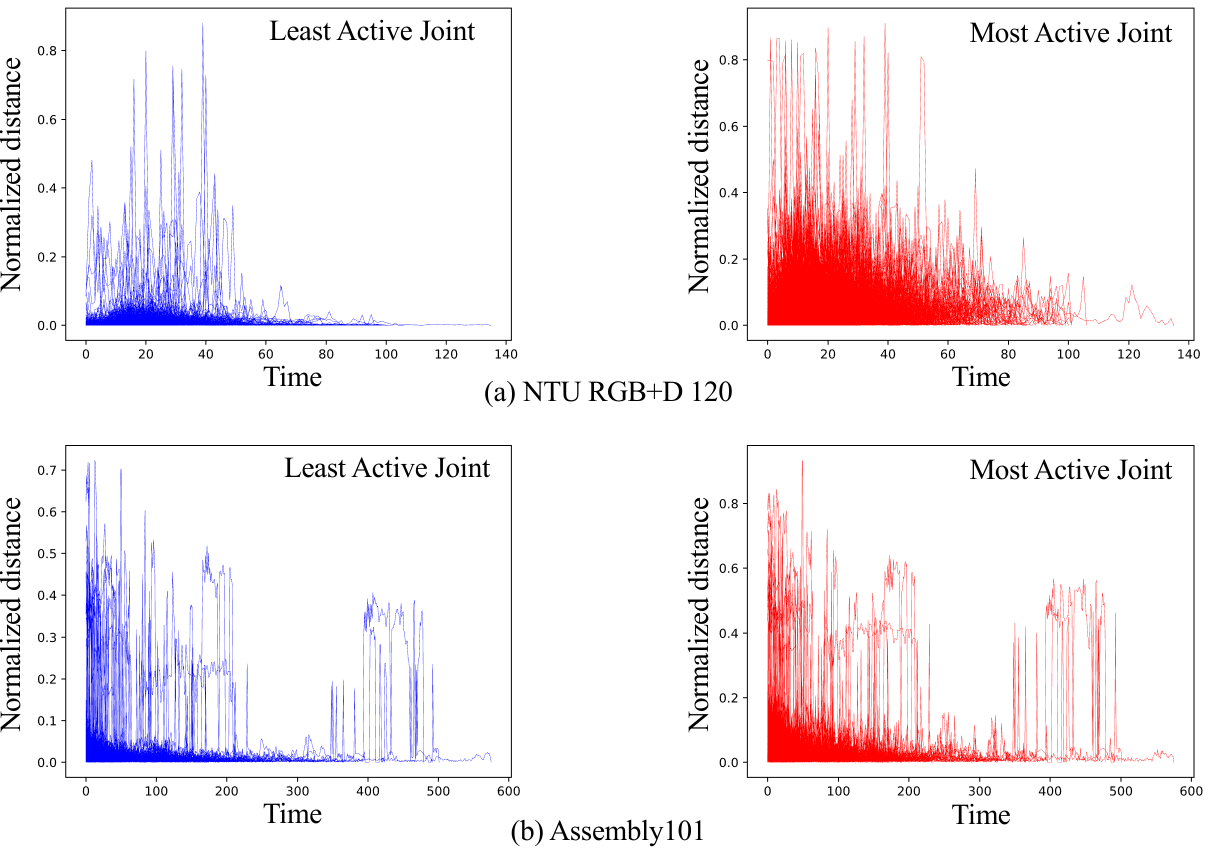}
    \caption{With a random pool of 1000 sequences, we observe that the least active joints can be viewed as static reference points, showing minimal movement in NTU RGB+D 120. In contrast, Assembly101 exhibits subtler distinctions between the most active and the least active joints. The Pearson correlation coefficient ($r$) between the distance values for these two joints yields a high value (0.93) for Assembly101, while $r=0.33$ for NTU RGB+D 120. These results suggest strong coupling among hand joints during motion, emphasizing the dominance of full-skeleton motion in hand poses. Our method leverages this understanding, balancing long-term motion patterns and short-term articulation changes by factorization.}
    \label{fig:ntu_assembly}
\end{figure*}

This section serves as an extension of~\cref{sec:skel_vs_hand} to statistically analyze the differences between hand poses and full-body poses. Action recognition datasets~\cite{shahroudy2016ntu, liu2019ntu}, which primarily focus on full-body actions, often include actions involving partial body movements. These actions exhibit limited global motion when viewed with respect to the entire skeleton, resulting in one or more relatively static joints. The change in the locations of moving joints with respect to such static joints can provide useful action cues. 
However, hand motions typically feature no such static reference points, as all the hand joints move together most of the time, making small changes in articulation to perform an action. To illustrate this difference, we randomly sample 1000 pose sequences from NTU RGB+D 120~\cite{liu2019ntu} (full-body) and Assembly101~\cite{sener2022assembly101} (hands). We take the distances covered between two consecutive frames for each joint to form a distance array for the corresponding joint $j$ in a given pose sequence, which is determined by-
\begin{equation}\label{eq:dist_arr}
d_j (t) = \big \lVert P_j(t) - P_j(t-1) \big \rVert 
\end{equation}
Here, $d_j (t)$ is the distance covered by joint $j$ at frame $t$ in reference to the previous frame, $P_j(t)$ and $P_j(t-1)$ are the 3D pose coordinates for joint $j$ at time $t$ and $t-1$, respectively. $ \lVert .  \rVert $ represents the Euclidean distance. Based on the sum of distances $D_j$ for each joint $j$, we define the \textit{\textbf{least active joint}} (static) and the \textit{\textbf{most active joint}} (dynamic) for a particular sequence using the following equations-
\begin{gather}
\begin{aligned}
    D_j =& \sum_{t=1}^{T} d_j(t) \\
    j_{sta} = \argmin_{j} D_j &\qquad j_{dyn} = \argmax_{j} D_j
\end{aligned}
\end{gather}
For each of the selected 1000 sequences, we take the two temporal sequences $\{d_{j_{sta}}(t)\}_{t=1}^T$ and $\{d_{j_{dyn}}(t)\}_{t=1}^T$, normalize the distance values using the diameters of the corresponding skeletons, and plot the sequences separately in~\cref{fig:ntu_assembly}. As can be observed, compared to the distances covered by the most active joints in NTU RGB+D 120, the least active joints show significantly lower movement, effectively serving as the static reference points. On the other hand, the distinction in distance arrays between the most and the least active joints in Assembly101 is less pronounced. In addition, we calculated the Pearson correlation coefficient, denoted as $r$, between $\{d_{j_{sta}}(t)\}_{t=1}^T$ and $\{d_{j_{dyn}}(t)\}_{t=1}^T$ for all Assembly101 sequences, resulting in a value of \textbf{0.93}. Conversely, for NTU RGB+D 120, the corresponding correlation coefficient is \textbf{0.33}. 
This suggests strong coupling among hand joints during motion, and full-skeleton movement is more dominant in hand poses compared to full-body poses. 
Consequently, modeling dependencies between spatiotemporally distant joints is less effective for the highly dynamic hand motion (also discussed in~\cref{sec:skel_vs_hand}). 
Therefore, by considering both long-term motion patterns and short-term articulation changes, our method facilitates efficient spatiotemporal factorization through micro-actions~(\cref{ss:form_ma}). We also incorporate the full-skeletal motion from the entire action during micro-action encoding, using a global wrist token as a reference~(\cref{ss:traj_enc}).

\section{2D vs. 3D Pose for Hand Actions}

For skeleton-based action recognition, PoseConv3D\cite{duan2022revisiting} proposes using 2D poses as input, arguing that the quality of pose estimation is superior in 2D. By constructing 3D heatmap volumes from 2D poses and employing a simple 3D-CNN, they surpass state-of-the-art GCN-based methods that rely on 3D poses. Incorporating CNN-based modeling for the pose stream also facilitates seamless integration with the RGB modality. In this section, we assess this proposition specifically within the context of hand skeletons.

\begin{figure}[!h]
    \centering
    \begin{subfigure}{\textwidth}
        \centering
        \includegraphics[width=0.95\textwidth,trim={0 7.5cm 7.5cm 0},clip]{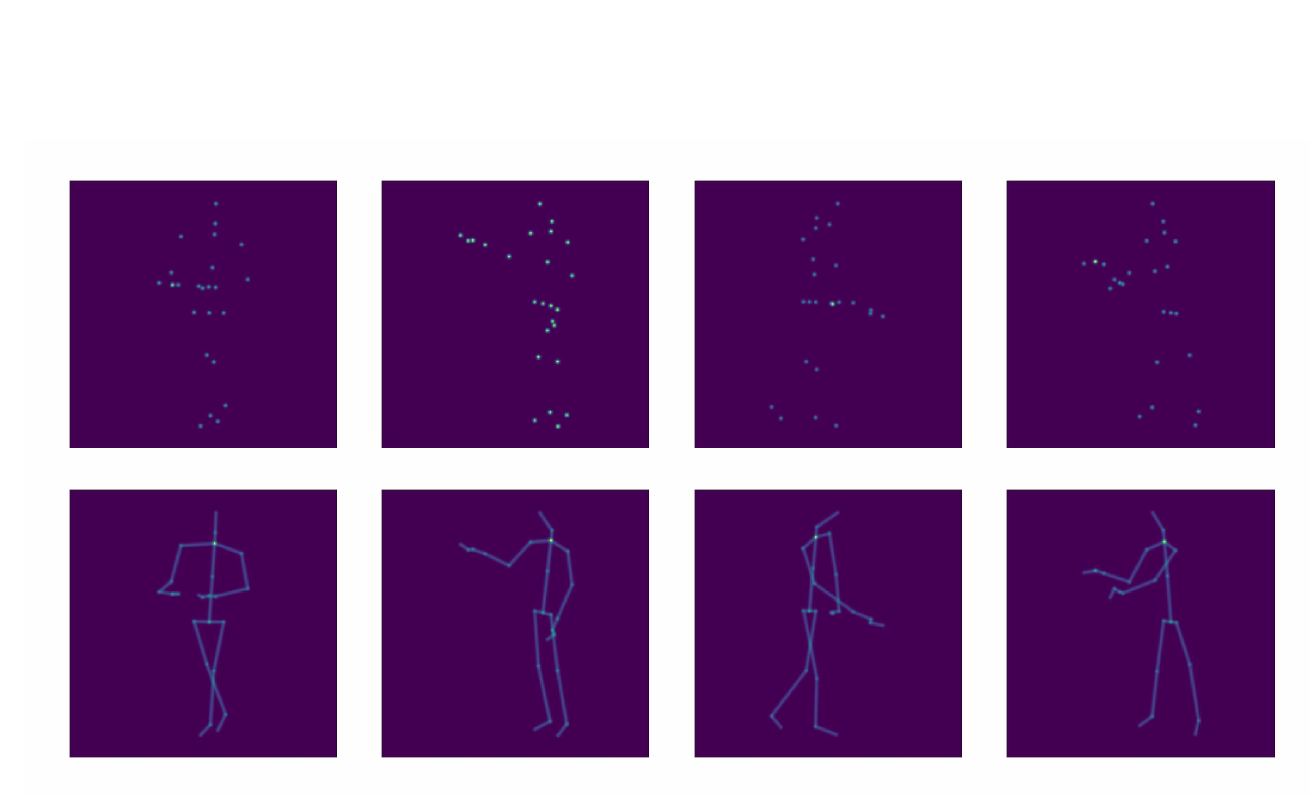}
        \caption{NTU RGB+D 120~\cite{liu2019ntu}.}
        \label{fig:ntu_heatmap}
    \end{subfigure}
    
    \begin{subfigure}{\textwidth}
        \centering
        \includegraphics[width=0.95\textwidth,trim={0 7.5cm 7.5cm 0},clip]{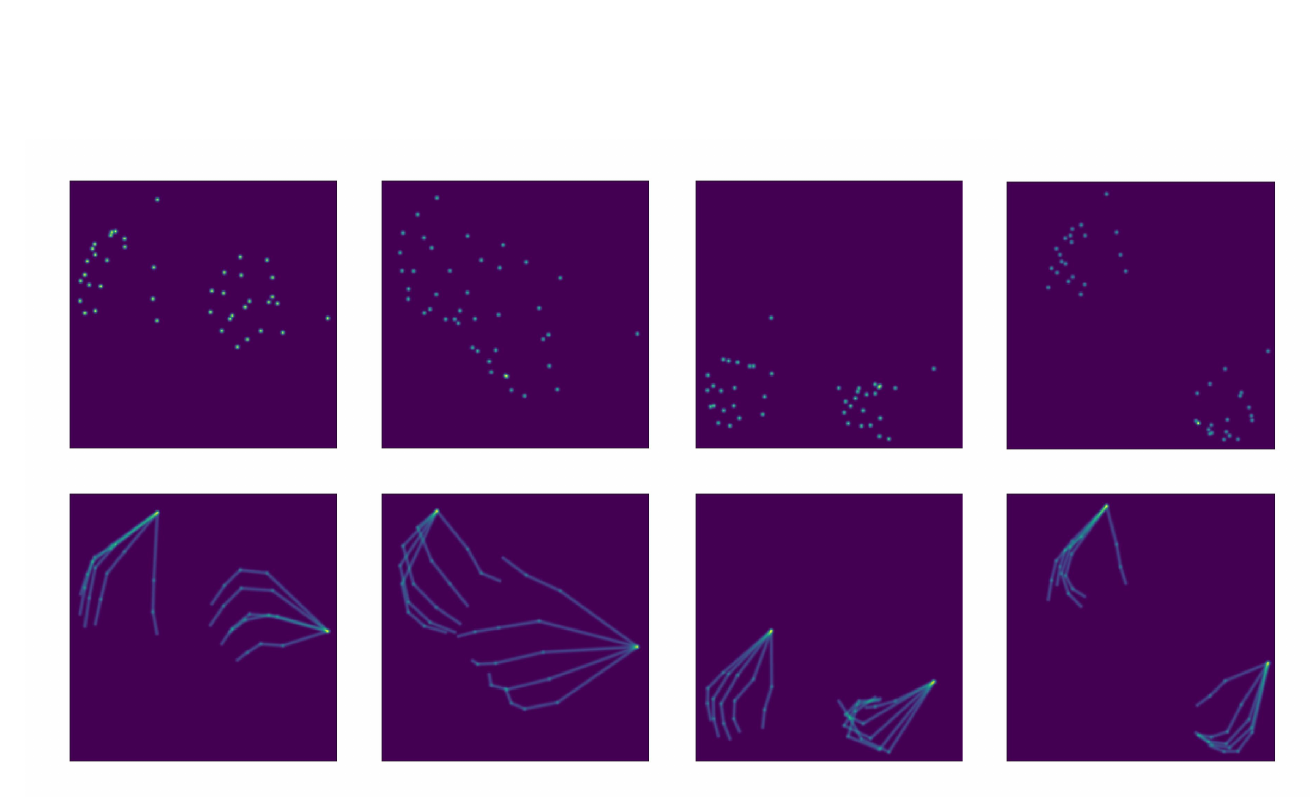}
        \caption{Assembly101~\cite{sener2022assembly101}.}
        \label{fig:assembly_heatmap}
    \end{subfigure}
    \caption{Heatmaps for joints and limbs for (a) full-body poses and (b) hand poses.}
    \label{fig:heatmaps_compare}
\end{figure}

\cref{fig:heatmaps_compare} illustrates sample heatmaps from NTU RGB+D 120~\cite{liu2019ntu} and Assembly101~\cite{sener2022assembly101}. Keypoints in full-body human poses are often prominently situated, with minimal self-occlusion, and the subject is typically centered within the frame. 
As viewed in~\cref{fig:ntu_heatmap}, reducing pose dimensions to 2D does not significantly compromise detail; rather, it enhances input reliability by simplifying the pose estimation problem. However, this advantage diminishes when applied to hand poses. Hand poses present unique challenges, such as frequent self-occlusion and closer proximity of the keypoints, which are exacerbated by reducing the dimension to 2D. 
To empirically analyze this phenomenon, we evaluate HandFormer-B/6 with 2D and 3D poses for recognizing verbs in Assembly101~\cite{sener2022assembly101} and report in~\cref{tab::2d_vs_3d}. 
This analysis reveals about $5$\% difference in favor of the 3D input.
Furthermore, PoseConv3D~\cite{duan2022revisiting} introduces a CNN-based approach with 2D keypoints, which directly utilizes heatmaps from the pose estimator or generates Gaussian heatmaps from the 2D coordinates. However, feeding heatmaps can diminish the clarity of keypoints to the model, particularly when they are in close proximity, as is often the case with hand poses. Hence, PoseConv3D~\cite{duan2022revisiting} performs poorly in recognizing hand actions, as evident in~\cref{tab::2d_vs_3d}.

\begin{table}[t]
  \centering
  \caption{\textbf{Impact of using 2D vs. 3D poses} as input for skeleton-based action recognition in hands. Experiments are done for verb recognition on Assembly101~\cite{sener2022assembly101}.}
  \begin{adjustbox}{width=0.7\linewidth}
  \begin{tabular}{@{}lcc@{}}
    \toprule
    \textbf{Method} & \textbf{\quad~Input Pose\quad} & \textbf{\quad~Verb Accuracy (\%)\quad} \\
    \midrule
    PoseConv3D~\cite{duan2022revisiting} & 2D & 46.71 \\ \midrule
    HandFormer-B/6 & 2D & 58.92 \\
    HandFormer-B/6 & 3D & 63.70 \\
    \bottomrule
  \end{tabular}
  \end{adjustbox}
  \label{tab::2d_vs_3d}
\end{table}

In summary, although skeleton-based methods represent a broader field for action recognition with poses, they often lack the necessary adaptation for directly addressing hand-specific actions.
This demands dedicated research on hand poses for hand-object interaction understanding.

\section{Alternatives for Frame Encoder}\label{ss:eff_frame_enc}
We utilize pre-trained ViT-g/14 and ViT-L/14 models from \mbox{DINOv2}~\cite{oquab2023dinov2} followed by a linear layer without any fine-tuning as the frame encoder $F$ for Assembly101~\cite{sener2022assembly101} and H2O~\cite{kwon2021h2o}, respectively. Except for our experiments on Assembly101~\cite{sener2022assembly101} with monochrome egocentric videos~(\cref{ss:ego}), all our multimodal results are obtained using DINOv2 features. This approach enables us to assess the effectiveness of image-based foundation models in videos, leveraging all-purpose features from RGB frames and achieving strong cross-view generalization performance~(\cref{ss:multiview}). Although this is our default choice offering easy adaptation to new datasets with faster training, it can be considered compute-heavy when deployed in a low-resource setting due to the large ViT backbones. For better efficiency during inference, we propose a pretraining scheme that allows us to use a ResNet50~\cite{he2016deep} that replaces the ViT without compromising accuracies, as shown in~\cref{tab:frame_enc}.
Specifically, we first train a TSM~\cite{lin2019tsm} model with a ResNet50~\cite{he2016deep} backbone for action recognition, utilizing all action clips and then dropping the classification layer.
This ResNet50 becomes the frozen image encoder in our proposed architecture, replacing ViT. During the training and inference of HandFormer, the TSM backbone operates as a true image model (ResNet50), as we employ it on individual frames without any channel shifting. The TSM features provided in the Assembly101~\cite{sener2022assembly101} are generated in this way, and we utilize them in our egocentric action recognition experiments~(\cref{ss:ego}). 

In~\cref{tab:frame_enc}, we present a comparison of the two backbone options for our frame encoder -- ResNet50 from TSM and ViT-g/14 from DINOv2. The ResNet50 outputs, enhanced through pretraining within TSM, incorporate domain-specific features and temporal encoding via channel shifting during training. As a result, the RGB-only variant achieves a 3\% higher action accuracy compared to using DINOv2 features alone. However, when introducing pose information, the image-based features are complemented by motion features, reducing the impact of motion understanding facilitated by the temporal shift mechanism of TSM in the ResNet50 encoder. Therefore, integrating pose data diminishes the pretraining advantage of ResNet50, resulting in a performance gap of less than 1\%.

\begin{table}[t]
  \centering
  \caption{\textbf{Comparison of different frame encoder options} on Assembly101~\cite{sener2022assembly101}. Frame-wise TSM features from pretrained ResNet50 perform better compared to all-purpose features generated by DINOv2 with a ViT-g/14 backbone. RGB-only variant greatly benefits from the pretraining as it works with domain-specific features for action recognition. However, incorporating complementary pose modality reduces the gain from pretraining.}
  \begin{adjustbox}{width=0.75\linewidth}
  \begin{tabular}{@{}llccc@{}}
    \toprule
    \textbf{Method Variant\quad\quad} & \textbf{Frame Encoder\quad\quad} &  \textbf{Action} & \textbf{Verb} & \textbf{Object}\\
    \midrule
     \multirow{2}{*}{RGB-only} & ViT-g/14     &  32.07 & 55.61 & 44.89 \\
                               & ResNet50     &  35.09 & 56.59 & 48.54 \\
    \midrule
     \multirow{2}{*}{Pose+RGB} & ViT-g/14     &  41.06 & 69.23 & 51.17 \\
                               & ResNet50     &  41.99 & 69.28 & 51.96 \\
    \bottomrule
  \end{tabular}
  \end{adjustbox}
  \label{tab:frame_enc}
\end{table}

\section{Maintaining High Temporal Resolution at Low Cost}
Our method is designed to perform action recognition efficiently in hand-object interaction videos. Obtaining efficiency in such a setup is challenging as we need to maintain a high temporal resolution to understand fine-grained hand movements that constitute the actions.
Therefore, we propose HandFormer using densely sampled pose frames and sparse RGB frames. In this section, we quantify the efficiency of this method compared to an alternative video model. As mentioned, understanding fine-grained hand motion demands a high temporal resolution to differentiate verb classes. For instance, relying on sparsely sampled frames may make actions like ``\textit{screwing}" and ``\textit{unscrewing}" indistinguishable. However, adopting a high temporal resolution with video models operating on RGB frames is challenging, primarily due to \textit{(i)} the excessive computation associated with performing spatiotemporal operations on numerous frames, and \textit{(ii)} the need to address redundancy in RGB frames to extract meaningful information. 

In \cref{tab:flops}, we compare the FLOPs of our model vs. an efficient video model, TSM~\cite{lin2019tsm} with a ResNet50 backbone when both maintain a high temporal resolution. The results reveal that our model operates at about $8\times$ fewer FLOPs. As TSM has a 2D backbone and no 3D convolutions, it is expected to represent the lower bound for the computational cost of a video model at that temporal resolution. For our frame encoder, we opt for the efficient alternative as described in~\cref{ss:eff_frame_enc}. The average duration of fine-grained actions in Assembly101~\cite{sener2022assembly101} is $1.7$ seconds. Following~\cite{sener2022assembly101}, we include an additional $0.5$ seconds of context on either side, resulting in an average of $2.7 \times 60 = 162$ frames per action clip. We use $K=8$ non-overlapping micro-actions, thus sampling 8 RGB frames and using the trajectory encoder eight times.

\begin{table}[t]
  \centering
  \caption{\textbf{Comparison of FLOPs between HandFormer and TSM}~\cite{lin2019tsm} when both maintain a high temporal resolution at 60~fps. The number of frames is determined by the average action duration in Assembly101~\cite{sener2022assembly101}, and we use eight non-overlapping micro-actions in our model.}
  \begin{adjustbox}{width=0.9\linewidth}
  \begin{tabular}{@{}lcccc@{}}
    \toprule
    \multirow{2}{*}{\textbf{Method}} & \multirow{2}{*}{\textbf{Component}} & \multirow{2}{*}{\textbf{\quad~GFLOPs\quad}} & \multirow{2}{*}{\textbf{Count}} & \textbf{Total} \\ 
    & & & & \textbf{\quad~GFLOPs\quad} \\
    \midrule
    TSM~\cite{lin2019tsm} & - & - & - & 669.79 \\
    \midrule    
    \multirow{5}{*}{HandFormer-B/21\quad} & Pose Estimator~\cite{han2020megatrack} & 0.30 & 162 & \multirow{5}{*}{84.01} \\ 
    & Frame Encoder & 4.12 & 8 & \\
    & Trajectory Encoder & 0.29 & 8 & \\ 
    & Multimodal Tokenizer & 0.01 & 8 & \\
    & Temporal Transformer & 0.05 & 1 & \\ 
    \bottomrule
  \end{tabular}
  \end{adjustbox}
  \label{tab:flops}
\end{table}

\section{Additional Details for Multimodal Training}
Our training recipe for the multimodal HandFormer involves initializing the trajectory encoder with pretrained weights and utilizing hand-object ROI crop within the frame encoder --- ensuring better use of pose and RGB, respectively.

\subsection{Pretraining Trajectory Encoder}
Encoding micro-action involves extracting RGB and pose features using frame encoder $F$ and trajectory encoder $T$, respectively. While the frame encoder stays frozen and provides the appearance features, the trajectory encoder is learned and is expected to capture the hand motion.
To effectively guide the trajectory encoder in achieving the desired encoding, we pretrain it for verb recognition solely using pose input. This approach leverages the inherent ability of pose data to capture hand motion, a key determinant of the verb while remaining agnostic to explicit information about interacting objects.
This pretraining scheme leads to a better initialization of the trajectory encoder in multimodal HandFormer for action recognition. In~\cref{tab:traj_pretrain}, we observe that initializing the trajectory encoder with pretrained weights leads to improved action recognition performance, particularly enhancing the recognition of verb classes.

\begin{table}[t]
  \centering
  \caption{\textbf{Initializing the trajectory encoder $T$ with pretrained weights} improves the overall performance with better verb recognition capability. Results are on Assembly101 \cite{sener2022assembly101} dataset. The initial weights for $T$ are obtained by training the model to predict the verb classes from pose-only input.}
  \begin{adjustbox}{width=0.75\linewidth}
  \begin{tabular}{@{}ccccc@{}}
    \toprule
    \multirow{2}{*}{\textbf{Frame Encoder}}& \textbf{\quad~Trajectory Encoder\quad} & \multicolumn{3}{c}{\textbf{Accuracy(\%)}}  \\ \cline{3-5}
    & \textbf{Pretraining} & \textbf{Action} & \textbf{Verb} & \textbf{Object}\\
    \midrule
    \multirow{2}{*}{ViT-g/14}& \xmark     &  39.79 & 67.40 & 50.69 \\
    & \cmark     &  41.06 & 69.23 & 51.17 \\
    \midrule
    \multirow{2}{*}{ResNet50} & \xmark     &  40.47 & 66.00 & 51.10 \\
    & \cmark     &  41.99 & 69.28 & 51.96 \\
    
    \bottomrule
  \end{tabular}
  \end{adjustbox}
  \label{tab:traj_pretrain}
\end{table}

\subsection{Hand-Object Interaction Crop}

\begin{table}[b]
  \centering
  \caption{\textbf{Ablation study comparing full vs. HOI cropped RGB frames} on Assembly101~\cite{sener2022assembly101}. Incorporating both full and cropped RGB frames allows for leveraging localized interaction details from hand crops and global contextual information from full frames, resulting in improved accuracy. HandFormer-B/21 is used with eight non-overlapping micro-actions.}
  \begin{adjustbox}{width=0.6\linewidth}
  \begin{tabular}{@{}ccccc@{}}
    \toprule
    \multirow{2}{*}{\textbf{Full Frame}}& \multirow{2}{*}{\textbf{\quad~HOI Crop\quad}} & \multicolumn{3}{c}{\textbf{Accuracy(\%)}}  \\ \cline{3-5}
    &  & \textbf{Action} & \textbf{Verb} & \textbf{Object}\\
    \midrule
    \cmark & \xmark     &  38.73 & 68.31 & 48.77 \\
    \xmark & \cmark     &  38.44 & 68.95 & 48.20 \\
    \cmark & \cmark     &  41.06 & 69.23 & 51.17 \\
    \bottomrule
  \end{tabular}
  \end{adjustbox}
  \label{tab:rgb_scale}
\end{table}

In hand-object interaction (HOI) videos, the region of interest typically centers around the hands, capturing crucial information about the interacting object and the type of interaction. Leveraging 3D hand poses obtained through a readily available pose estimator~\cite{han2020megatrack}, we project these poses onto RGB frames, extract the enclosing rectangle of the projected 2D pose, and expand it by 25\% to define the ROI crop. 
However, relying solely on the cropped region can occasionally mislead the model for three potential reasons: \textit{i)} failure of the pose estimator on certain frames, leading to the absence of useful features from the RGB frames, \textit{ii)} the full object might not be visible when the crop is taken based on hand poses only, and \textit{iii)} hand crops have limitations in capturing global changes compared to the full frames. Hence, to capitalize on both the localized interaction information of hand crops and the global contextual information provided by full frames, our model combines them both. If a valid hand crop is found, we take the full and cropped RGB frames, pass them through the frame encoder, average their features, and re-normalize them to unit norm. This full vs. HOI crop ablation is shown in \cref{tab:rgb_scale}, in which combining both performs better than the alternatives.

\section{Efficiency Comparison with Shift-GCN}
While MS-G3D~\cite{liu2020disentangling} and ISTA-Net~\cite{wen2023interactive} show state-of-the-art performance for action recognition with hand poses, they are not efficiency-focused. Our HandFormer outperforms them with significantly fewer FLOPs. However, HandFormer-B/6 prioritizes efficiency while slightly trading off accuracy. Therefore, we implement and test an efficiency-focused baseline, ShiftGCN~\cite{cheng2020skeleton}, for verb recognition on Assembly101~\cite{sener2022assembly101} and compare it to HandFormer-B/6 in~\cref{tab::shift_gcn}. While Shift-GCN relies on graph shift operations and pointwise convolutions for efficiency, our model identifies the crucial joints, \ie, the fingertips and the wrist joint, and processes only these joints to reduce FLOPs substantially. As evident from~\cref{tab::shift_gcn}, our model outperforms Shift-GCN while incurring lower FLOPs.

\begin{table}[h]
  \centering
  \caption{\textbf{Comparison of HandFormer-B/6 with Shift-GCN}, an efficiency-focused baseline for skeleton-based action recognition. Experiments are done for verb recognition on Assembly101~\cite{sener2022assembly101}.}
  \begin{adjustbox}{width=0.75\linewidth}
  \begin{tabular}{@{}lcc@{}}
    \toprule
    \textbf{Method} & \textbf{\quad~GFLOPs\quad} & \textbf{\quad~Verb Accuracy (\%)\quad} \\
    \midrule
    Shift-GCN~\cite{cheng2020skeleton} & 2.11 & 63.14 \\
    HandFormer-B/6 & 1.33 & 63.70 \\
    \bottomrule
  \end{tabular}
  \end{adjustbox}
  \label{tab::shift_gcn}
\end{table}

\begin{figure*}[!h]
 \centering
 \includegraphics[width=\textwidth]
 {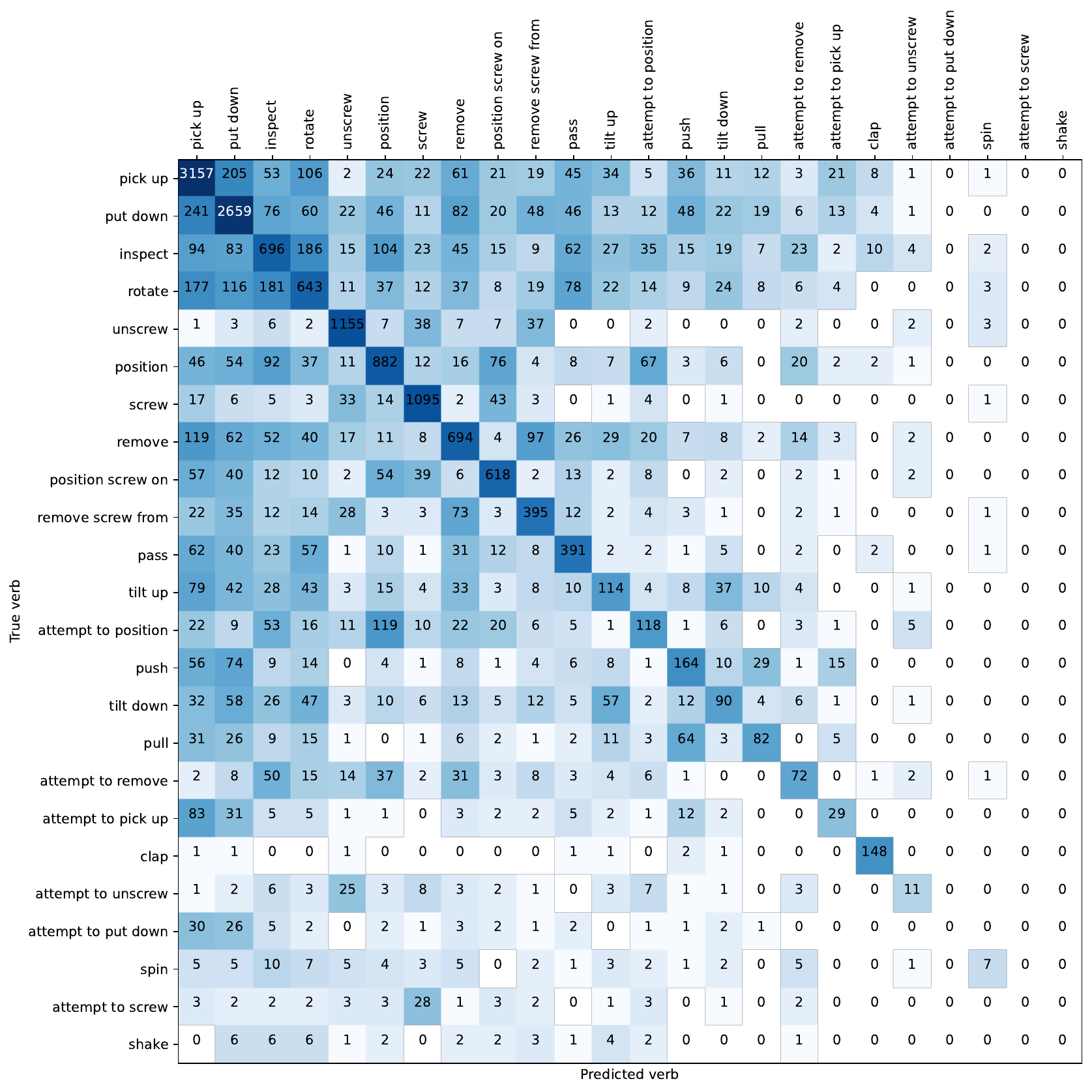}
 \caption{Confusion matrix for pose-only verb recognition with HandFormer-L/21.}
 \label{fig:conf_mat}
\end{figure*}

\section{Qualitative Analysis}
In this section, we analyze the class-wise verb accuracy using the pose-only HandFormer, aiming to identify the model's limitations. Furthermore, we examine the multimodal aspect of action recognition and its role in alleviating object misclassification.

\subsection{Pose-only Performance}
\cref{fig:conf_mat} displays the confusion matrix for verb classes using HandFormer-L/21 on the test set. Notably, \textit{inspect}, \textit{rotate}, \textit{position}, and \textit{remove} verbs present recognition challenges despite ample dataset samples. One potential explanation for this phenomenon is the shared presence of certain signature movements among these classes, which also occur in two head classes, namely, \textit{pick up} and \textit{put down}. Another interesting observation in the results is the frequent classification of \textit{`attempt to x'} classes as \textit{`x'}. This is expected, as determining the successful completion of a task adds another layer of complexity to these classes, especially when relying solely on pose information without considering changes in the appearance of the interacting object throughout the clip.

\begin{table}[t]
    \centering
    \caption{Action classification by our model with and without sampling an RGB frame. Incorrect predictions are highlighted in \textcolor{red}{red}, while correct predictions are marked in \textcolor{darkpastelgreen}{green}.}
    \begin{tabular}{lccc}
    \hline
    \multirow{2}{*}{\emph{Pose + 0 RGB}} & \textcolor{darkpastelgreen}{Put down} &  \textcolor{darkpastelgreen}{Put down} & \textcolor{darkpastelgreen}{Put down} \\
      & \textcolor{red}{screwdriver} & \textcolor{red}{screw} & \textcolor{red}{partial toy} \\
     
    \hline
    RGB Sample & \adjustbox{valign=c}{\includegraphics[width=0.18\textwidth]{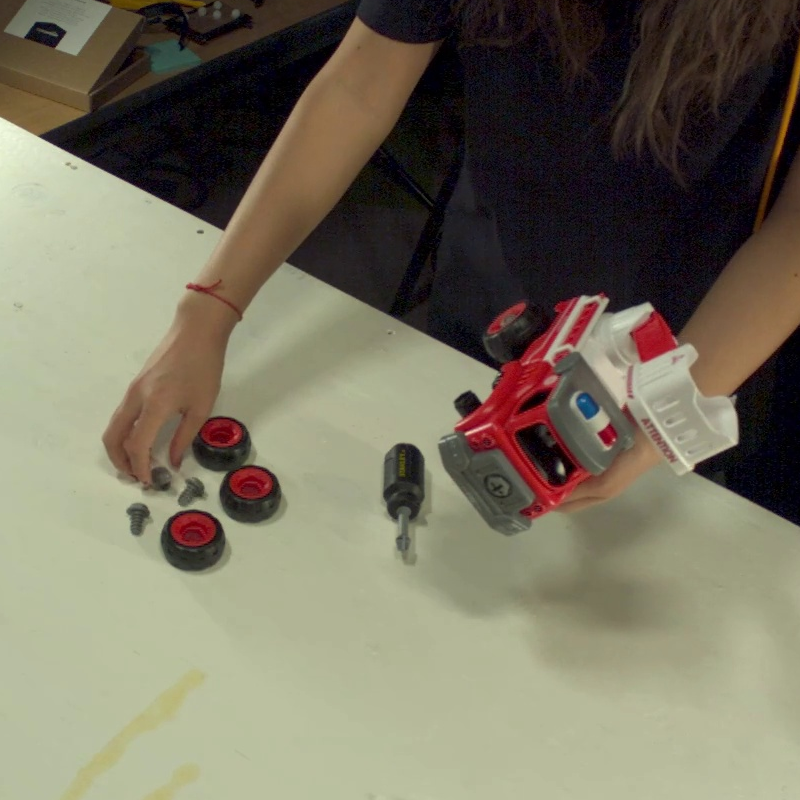}}  & \adjustbox{valign=c}{\includegraphics[width=0.18\textwidth]{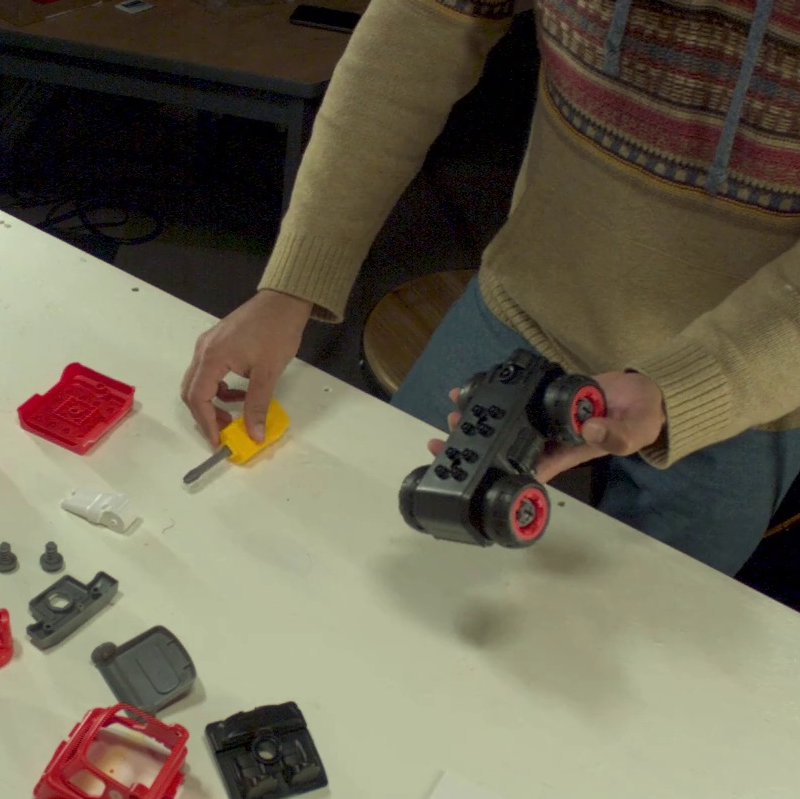}} & \adjustbox{valign=c}{\includegraphics[width=0.18\textwidth]{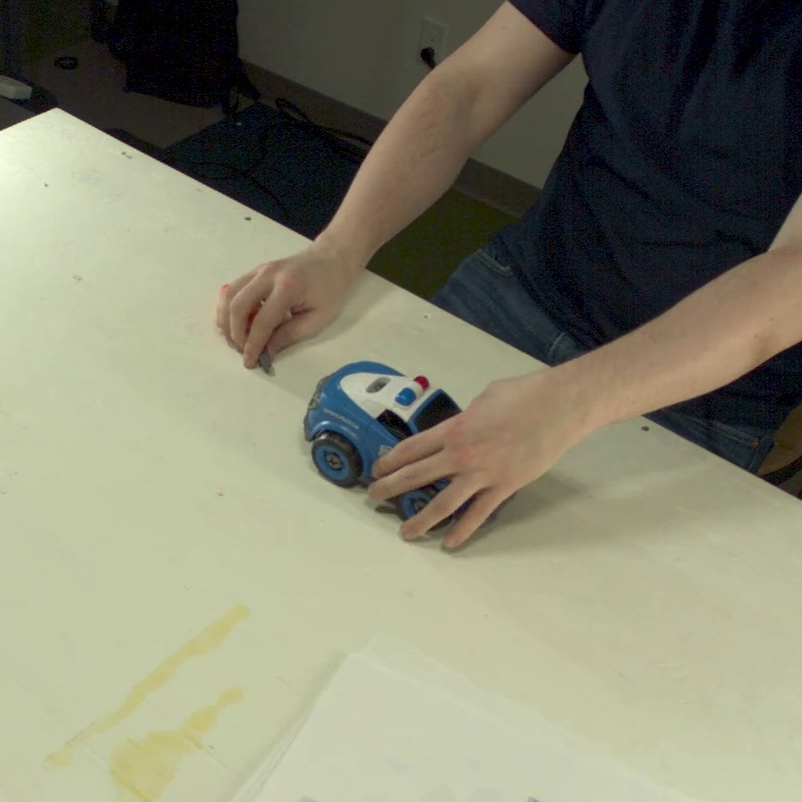}} \\
    \hline
    
    \multirow{2}{*}{\emph{Pose + 1 RGB}} & \textcolor{darkpastelgreen}{Put down} & \textcolor{darkpastelgreen}{Put down} & \textcolor{darkpastelgreen}{Put down} \\
     & \textcolor{darkpastelgreen}{screw}  & \textcolor{darkpastelgreen}{screwdriver} & \textcolor{darkpastelgreen}{finished toy} \\
    \hline
    \end{tabular}
    \label{tab:rgb_added}
\end{table}

\subsection{Multimodal Fusion}
To gain insights into how appearance information from RGB complements pose-based models in hand-object interaction scenarios, we analyze samples involving \textit{put down} actions. In~\cref{tab:rgb_added}, we showcase the action classes predicted for these samples using our pose-only model, referred to as \emph{Pose + 0 RGB}. In these samples, the model successfully detected the verb but struggled with object classification. This challenge arises due to similarities in articulations observed during tasks such as grasping a screwdriver and a screw or differentiating between a partially assembled toy and a completed one. These similarities lead to misclassifications by the pose-only model. However, introducing a single RGB frame, denoted as \emph{Pose + 1 RGB}, enhances the model's ability to correctly identify the relevant object by providing visual context. This observation highlights the limitations of recognizing actions, \ie verb+object, solely from hand poses, emphasizing the importance of incorporating visual cues.

\bibliographystyle{splncs04}
\bibliography{main}

\begin{thebibliography}{10}
\providecommand{\url}[1]{\texttt{#1}}
\providecommand{\urlprefix}{URL }
\providecommand{\doi}[1]{https://doi.org/#1}

\bibitem{ahn2023star}
Ahn, D., Kim, S., Hong, H., Ko, B.C.: Star-transformer: a spatio-temporal cross attention transformer for human action recognition. In: Proceedings of the IEEE/CVF Winter Conference on Applications of Computer Vision. pp. 3330--3339 (2023)

\bibitem{arnab2021vivit}
Arnab, A., Dehghani, M., Heigold, G., Sun, C., Lu{\v{c}}i{\'c}, M., Schmid, C.: Vivit: A video vision transformer. In: Proceedings of the IEEE/CVF international conference on computer vision. pp. 6836--6846 (2021)

\bibitem{bertasius2021space}
Bertasius, G., Wang, H., Torresani, L.: Is space-time attention all you need for video understanding? In: ICML. vol.~2, p.~4 (2021)

\bibitem{bolya2022token}
Bolya, D., Fu, C.Y., Dai, X., Zhang, P., Feichtenhofer, C., Hoffman, J.: Token merging: Your vit but faster. arXiv preprint arXiv:2210.09461  (2022)

\bibitem{bruce2021multimodal}
Bruce, X., Liu, Y., Chan, K.C.: Multimodal fusion via teacher-student network for indoor action recognition. In: Proceedings of the AAAI Conference on Artificial Intelligence. vol.~35, pp. 3199--3207 (2021)

\bibitem{bruce2022mmnet}
Bruce, X., Liu, Y., Zhang, X., Zhong, S.h., Chan, K.C.: Mmnet: A model-based multimodal network for human action recognition in rgb-d videos. IEEE Transactions on Pattern Analysis and Machine Intelligence  \textbf{45}(3),  3522--3538 (2022)

\bibitem{caetano2019skelemotion}
Caetano, C., Sena, J., Br{\'e}mond, F., Dos~Santos, J.A., Schwartz, W.R.: Skelemotion: A new representation of skeleton joint sequences based on motion information for 3d action recognition. In: 2019 16th IEEE international conference on advanced video and signal based surveillance (AVSS). pp.~1--8. IEEE (2019)

\bibitem{cao2017body}
Cao, C., Zhang, Y., Zhang, C., Lu, H.: Body joint guided 3-d deep convolutional descriptors for action recognition. IEEE transactions on cybernetics  \textbf{48}(3),  1095--1108 (2017)

\bibitem{carreira2017quo}
Carreira, J., Zisserman, A.: Quo vadis, action recognition? a new model and the kinetics dataset. In: proceedings of the IEEE Conference on Computer Vision and Pattern Recognition. pp. 6299--6308 (2017)

\bibitem{chatterjee2023opening}
Chatterjee, D., Sener, F., Ma, S., Yao, A.: Opening the vocabulary of egocentric actions. In: Thirty-seventh Conference on Neural Information Processing Systems (2023)

\bibitem{cheng2020skeleton}
Cheng, K., Zhang, Y., He, X., Chen, W., Cheng, J., Lu, H.: Skeleton-based action recognition with shift graph convolutional network. In: Proceedings of the IEEE/CVF conference on computer vision and pattern recognition. pp. 183--192 (2020)

\bibitem{cheron2015p}
Ch{\'e}ron, G., Laptev, I., Schmid, C.: P-cnn: Pose-based cnn features for action recognition. In: Proceedings of the IEEE international conference on computer vision. pp. 3218--3226 (2015)

\bibitem{cho2023transformer}
Cho, H., Kim, C., Kim, J., Lee, S., Ismayilzada, E., Baek, S.: Transformer-based unified recognition of two hands manipulating objects. In: Proceedings of the IEEE/CVF Conference on Computer Vision and Pattern Recognition. pp. 4769--4778 (2023)

\bibitem{damen2018scaling}
Damen, D., Doughty, H., Farinella, G.M., Fidler, S., Furnari, A., Kazakos, E., Moltisanti, D., Munro, J., Perrett, T., Price, W., et~al.: Scaling egocentric vision: The epic-kitchens dataset. In: Proceedings of the European conference on computer vision (ECCV). pp. 720--736 (2018)

\bibitem{das2021vpn++}
Das, S., Dai, R., Yang, D., Bremond, F.: Vpn++: Rethinking video-pose embeddings for understanding activities of daily living. IEEE Transactions on Pattern Analysis and Machine Intelligence  \textbf{44}(12),  9703--9717 (2021)

\bibitem{das2020vpn}
Das, S., Sharma, S., Dai, R., Bremond, F., Thonnat, M.: Vpn: Learning video-pose embedding for activities of daily living. In: Computer Vision--ECCV 2020: 16th European Conference, Glasgow, UK, August 23--28, 2020, Proceedings, Part IX 16. pp. 72--90. Springer (2020)

\bibitem{devlin2018bert}
Devlin, J., Chang, M.W., Lee, K., Toutanova, K.: Bert: Pre-training of deep bidirectional transformers for language understanding. arXiv preprint arXiv:1810.04805  (2018)

\bibitem{dosovitskiy2020image}
Dosovitskiy, A., Beyer, L., Kolesnikov, A., Weissenborn, D., Zhai, X., Unterthiner, T., Dehghani, M., Minderer, M., Heigold, G., Gelly, S., et~al.: An image is worth 16x16 words: Transformers for image recognition at scale. arXiv preprint arXiv:2010.11929  (2020)

\bibitem{du2015hierarchical}
Du, Y., Wang, W., Wang, L.: Hierarchical recurrent neural network for skeleton based action recognition. In: Proceedings of the IEEE conference on computer vision and pattern recognition. pp. 1110--1118 (2015)

\bibitem{duan2022revisiting}
Duan, H., Zhao, Y., Chen, K., Lin, D., Dai, B.: Revisiting skeleton-based action recognition. In: Proceedings of the IEEE/CVF Conference on Computer Vision and Pattern Recognition. pp. 2969--2978 (2022)

\bibitem{fayyaz2022adaptive}
Fayyaz, M., Koohpayegani, S.A., Jafari, F.R., Sengupta, S., Joze, H.R.V., Sommerlade, E., Pirsiavash, H., Gall, J.: Adaptive token sampling for efficient vision transformers. In: European Conference on Computer Vision. pp. 396--414. Springer (2022)

\bibitem{feichtenhofer2020x3d}
Feichtenhofer, C.: X3d: Expanding architectures for efficient video recognition. In: Proceedings of the IEEE/CVF conference on computer vision and pattern recognition. pp. 203--213 (2020)

\bibitem{feichtenhofer2019slowfast}
Feichtenhofer, C., Fan, H., Malik, J., He, K.: Slowfast networks for video recognition. In: Proceedings of the IEEE/CVF international conference on computer vision. pp. 6202--6211 (2019)

\bibitem{gan2023keyframe}
Gan, M., Liu, J., He, Y., Chen, A., Ma, Q.: Keyframe selection via deep reinforcement learning for skeleton-based gesture recognition. IEEE Robotics and Automation Letters  (2023)

\bibitem{girdhar2021anticipative}
Girdhar, R., Grauman, K.: Anticipative video transformer. In: Proceedings of the IEEE/CVF international conference on computer vision. pp. 13505--13515 (2021)

\bibitem{grauman2022ego4d}
Grauman, K., Westbury, A., Byrne, E., Chavis, Z., Furnari, A., Girdhar, R., Hamburger, J., Jiang, H., Liu, M., Liu, X., et~al.: Ego4d: Around the world in 3,000 hours of egocentric video. In: Proceedings of the IEEE/CVF Conference on Computer Vision and Pattern Recognition. pp. 18995--19012 (2022)

\bibitem{han2020megatrack}
Han, S., Liu, B., Cabezas, R., Twigg, C.D., Zhang, P., Petkau, J., Yu, T.H., Tai, C.J., Akbay, M., Wang, Z., et~al.: Megatrack: monochrome egocentric articulated hand-tracking for virtual reality. ACM Transactions on Graphics (ToG)  \textbf{39}(4),  87--1 (2020)

\bibitem{han2022umetrack}
Han, S., Wu, P.c., Zhang, Y., Liu, B., Zhang, L., Wang, Z., Si, W., Zhang, P., Cai, Y., Hodan, T., et~al.: Umetrack: Unified multi-view end-to-end hand tracking for vr. In: SIGGRAPH Asia 2022 Conference Papers. pp.~1--9 (2022)

\bibitem{he2016deep}
He, K., Zhang, X., Ren, S., Sun, J.: Deep residual learning for image recognition. In: Proceedings of the IEEE conference on computer vision and pattern recognition. pp. 770--778 (2016)

\bibitem{hou2018spatial}
Hou, J., Wang, G., Chen, X., Xue, J.H., Zhu, R., Yang, H.: Spatial-temporal attention res-tcn for skeleton-based dynamic hand gesture recognition. In: Proceedings of the European conference on computer vision (ECCV) workshops. pp.~0--0 (2018)

\bibitem{hou2016skeleton}
Hou, Y., Li, Z., Wang, P., Li, W.: Skeleton optical spectra-based action recognition using convolutional neural networks. IEEE Transactions on Circuits and Systems for Video Technology  \textbf{28}(3),  807--811 (2016)

\bibitem{jaegle2021perceiver}
Jaegle, A., Gimeno, F., Brock, A., Vinyals, O., Zisserman, A., Carreira, J.: Perceiver: General perception with iterative attention. In: International conference on machine learning. pp. 4651--4664. PMLR (2021)

\bibitem{kazakos2019epic}
Kazakos, E., Nagrani, A., Zisserman, A., Damen, D.: Epic-fusion: Audio-visual temporal binding for egocentric action recognition. In: Proceedings of the IEEE/CVF International Conference on Computer Vision. pp. 5492--5501 (2019)

\bibitem{kwon2021h2o}
Kwon, T., Tekin, B., St{\"u}hmer, J., Bogo, F., Pollefeys, M.: H2o: Two hands manipulating objects for first person interaction recognition. In: Proceedings of the IEEE/CVF International Conference on Computer Vision. pp. 10138--10148 (2021)

\bibitem{lea2017temporal}
Lea, C., Flynn, M.D., Vidal, R., Reiter, A., Hager, G.D.: Temporal convolutional networks for action segmentation and detection. In: proceedings of the IEEE Conference on Computer Vision and Pattern Recognition. pp. 156--165 (2017)

\bibitem{lee2024echowrist}
Lee, C.J., Zhang, R., Agarwal, D., Yu, T.C., Gunda, V., Lopez, O., Kim, J., Yin, S., Deng, B., Li, K., et~al.: Echowrist: Continuous hand pose tracking and hand-object interaction recognition using low-power active acoustic sensing on a wristband. arXiv preprint arXiv:2401.17409  (2024)

\bibitem{li2021two}
Li, C., Li, S., Gao, Y., Zhang, X., Li, W.: A two-stream neural network for pose-based hand gesture recognition. IEEE Transactions on Cognitive and Developmental Systems  \textbf{14}(4),  1594--1603 (2021)

\bibitem{li2020sgm}
Li, J., Xie, X., Pan, Q., Cao, Y., Zhao, Z., Shi, G.: Sgm-net: Skeleton-guided multimodal network for action recognition. Pattern Recognition  \textbf{104},  107356 (2020)

\bibitem{li2019actional}
Li, M., Chen, S., Chen, X., Zhang, Y., Wang, Y., Tian, Q.: Actional-structural graph convolutional networks for skeleton-based action recognition. In: Proceedings of the IEEE/CVF conference on computer vision and pattern recognition. pp. 3595--3603 (2019)

\bibitem{lin2019tsm}
Lin, J., Gan, C., Han, S.: Tsm: Temporal shift module for efficient video understanding. In: Proceedings of the IEEE/CVF international conference on computer vision. pp. 7083--7093 (2019)

\bibitem{liu2019ntu}
Liu, J., Shahroudy, A., Perez, M., Wang, G., Duan, L.Y., Kot, A.C.: Ntu rgb+ d 120: A large-scale benchmark for 3d human activity understanding. IEEE transactions on pattern analysis and machine intelligence  \textbf{42}(10),  2684--2701 (2019)

\bibitem{liu2021neuropose}
Liu, Y., Zhang, S., Gowda, M.: Neuropose: 3d hand pose tracking using emg wearables. In: Proceedings of the Web Conference 2021. pp. 1471--1482 (2021)

\bibitem{liu2022video}
Liu, Z., Ning, J., Cao, Y., Wei, Y., Zhang, Z., Lin, S., Hu, H.: Video swin transformer. In: Proceedings of the IEEE/CVF conference on computer vision and pattern recognition. pp. 3202--3211 (2022)

\bibitem{liu2020disentangling}
Liu, Z., Zhang, H., Chen, Z., Wang, Z., Ouyang, W.: Disentangling and unifying graph convolutions for skeleton-based action recognition. In: Proceedings of the IEEE/CVF conference on computer vision and pattern recognition. pp. 143--152 (2020)

\bibitem{ma2022hand}
Ma, J., Damen, D.: Hand-object interaction reasoning. In: 2022 18th IEEE International Conference on Advanced Video and Signal Based Surveillance (AVSS). pp.~1--8. IEEE (2022)

\bibitem{ohkawa2023assemblyhands}
Ohkawa, T., He, K., Sener, F., Hodan, T., Tran, L., Keskin, C.: Assemblyhands: Towards egocentric activity understanding via 3d hand pose estimation. In: Proceedings of the IEEE/CVF Conference on Computer Vision and Pattern Recognition. pp. 12999--13008 (2023)

\bibitem{oquab2023dinov2}
Oquab, M., Darcet, T., Moutakanni, T., Vo, H., Szafraniec, M., Khalidov, V., Fernandez, P., Haziza, D., Massa, F., El-Nouby, A., et~al.: Dinov2: Learning robust visual features without supervision. arXiv preprint arXiv:2304.07193  (2023)

\bibitem{patrick2021keeping}
Patrick, M., Campbell, D., Asano, Y., Misra, I., Metze, F., Feichtenhofer, C., Vedaldi, A., Henriques, J.F.: Keeping your eye on the ball: Trajectory attention in video transformers. Advances in neural information processing systems  \textbf{34},  12493--12506 (2021)

\bibitem{plizzari2021spatial}
Plizzari, C., Cannici, M., Matteucci, M.: Spatial temporal transformer network for skeleton-based action recognition. In: Pattern Recognition. ICPR International Workshops and Challenges: Virtual Event, January 10--15, 2021, Proceedings, Part III. pp. 694--701. Springer (2021)

\bibitem{rajasegaran2023benefits}
Rajasegaran, J., Pavlakos, G., Kanazawa, A., Feichtenhofer, C., Malik, J.: On the benefits of 3d pose and tracking for human action recognition. In: Proceedings of the IEEE/CVF Conference on Computer Vision and Pattern Recognition. pp. 640--649 (2023)

\bibitem{sabater2021domain}
Sabater, A., Alonso, I., Montesano, L., Murillo, A.C.: Domain and view-point agnostic hand action recognition. IEEE Robotics and Automation Letters  \textbf{6}(4),  7823--7830 (2021)

\bibitem{sener2022assembly101}
Sener, F., Chatterjee, D., Shelepov, D., He, K., Singhania, D., Wang, R., Yao, A.: Assembly101: A large-scale multi-view video dataset for understanding procedural activities. In: Proceedings of the IEEE/CVF Conference on Computer Vision and Pattern Recognition. pp. 21096--21106 (2022)

\bibitem{shahroudy2016ntu}
Shahroudy, A., Liu, J., Ng, T.T., Wang, G.: Ntu rgb+ d: A large scale dataset for 3d human activity analysis. In: Proceedings of the IEEE conference on computer vision and pattern recognition. pp. 1010--1019 (2016)

\bibitem{shan2020understanding}
Shan, D., Geng, J., Shu, M., Fouhey, D.F.: Understanding human hands in contact at internet scale. In: Proceedings of the IEEE/CVF conference on computer vision and pattern recognition. pp. 9869--9878 (2020)

\bibitem{shi2019two}
Shi, L., Zhang, Y., Cheng, J., Lu, H.: Two-stream adaptive graph convolutional networks for skeleton-based action recognition. In: Proceedings of the IEEE/CVF conference on computer vision and pattern recognition. pp. 12026--12035 (2019)

\bibitem{simonyan2014two}
Simonyan, K., Zisserman, A.: Two-stream convolutional networks for action recognition in videos. Advances in neural information processing systems  \textbf{27} (2014)

\bibitem{soo2017interpretable}
Soo~Kim, T., Reiter, A.: Interpretable 3d human action analysis with temporal convolutional networks. In: Proceedings of the IEEE conference on computer vision and pattern recognition workshops. pp. 20--28 (2017)

\bibitem{tran2015learning}
Tran, D., Bourdev, L., Fergus, R., Torresani, L., Paluri, M.: Learning spatiotemporal features with 3d convolutional networks. In: Proceedings of the IEEE international conference on computer vision. pp. 4489--4497 (2015)

\bibitem{tran2018closer}
Tran, D., Wang, H., Torresani, L., Ray, J., LeCun, Y., Paluri, M.: A closer look at spatiotemporal convolutions for action recognition. In: Proceedings of the IEEE conference on Computer Vision and Pattern Recognition. pp. 6450--6459 (2018)

\bibitem{vaswani2017attention}
Vaswani, A., Shazeer, N., Parmar, N., Uszkoreit, J., Jones, L., Gomez, A.N., Kaiser, {\L}., Polosukhin, I.: Attention is all you need. Advances in neural information processing systems  \textbf{30} (2017)

\bibitem{vemulapalli2014human}
Vemulapalli, R., Arrate, F., Chellappa, R.: Human action recognition by representing 3d skeletons as points in a lie group. In: Proceedings of the IEEE conference on computer vision and pattern recognition. pp. 588--595 (2014)

\bibitem{vondrick2016anticipating}
Vondrick, C., Pirsiavash, H., Torralba, A.: Anticipating visual representations from unlabeled video. In: Proceedings of the IEEE conference on computer vision and pattern recognition. pp. 98--106 (2016)

\bibitem{wang2012mining}
Wang, J., Liu, Z., Wu, Y., Yuan, J.: Mining actionlet ensemble for action recognition with depth cameras. In: 2012 IEEE conference on computer vision and pattern recognition. pp. 1290--1297. IEEE (2012)

\bibitem{weiyao2021fusion}
Weiyao, X., Muqing, W., Min, Z., Ting, X.: Fusion of skeleton and rgb features for rgb-d human action recognition. IEEE Sensors Journal  \textbf{21}(17),  19157--19164 (2021)

\bibitem{wen2023hierarchical}
Wen, Y., Pan, H., Yang, L., Pan, J., Komura, T., Wang, W.: Hierarchical temporal transformer for 3d hand pose estimation and action recognition from egocentric rgb videos. In: Proceedings of the IEEE/CVF Conference on Computer Vision and Pattern Recognition. pp. 21243--21253 (2023)

\bibitem{wen2023interactive}
Wen, Y., Tang, Z., Pang, Y., Ding, B., Liu, M.: Interactive spatiotemporal token attention network for skeleton-based general interactive action recognition. arXiv preprint arXiv:2307.07469  (2023)

\bibitem{wu2022memvit}
Wu, C.Y., Li, Y., Mangalam, K., Fan, H., Xiong, B., Malik, J., Feichtenhofer, C.: Memvit: Memory-augmented multiscale vision transformer for efficient long-term video recognition. In: Proceedings of the IEEE/CVF Conference on Computer Vision and Pattern Recognition. pp. 13587--13597 (2022)

\bibitem{xie2018rethinking}
Xie, S., Sun, C., Huang, J., Tu, Z., Murphy, K.: Rethinking spatiotemporal feature learning: Speed-accuracy trade-offs in video classification. In: Proceedings of the European conference on computer vision (ECCV). pp. 305--321 (2018)

\bibitem{yan2018spatial}
Yan, S., Xiong, Y., Lin, D.: Spatial temporal graph convolutional networks for skeleton-based action recognition. In: Proceedings of the AAAI conference on artificial intelligence. vol.~32 (2018)

\bibitem{zhang2017geometric}
Zhang, S., Liu, X., Xiao, J.: On geometric features for skeleton-based action recognition using multilayer lstm networks. In: 2017 IEEE winter conference on applications of computer vision (WACV). pp. 148--157. IEEE (2017)

\bibitem{zhang2019graph}
Zhang, X., Xu, C., Tian, X., Tao, D.: Graph edge convolutional neural networks for skeleton-based action recognition. IEEE transactions on neural networks and learning systems  \textbf{31}(8),  3047--3060 (2019)

\bibitem{zhang2021stst}
Zhang, Y., Wu, B., Li, W., Duan, L., Gan, C.: Stst: Spatial-temporal specialized transformer for skeleton-based action recognition. In: Proceedings of the 29th ACM International Conference on Multimedia. pp. 3229--3237 (2021)

\end{thebibliography}
\end{document}